\documentclass{article}
\usepackage{esb_preprint,times}
\usepackage{amsmath,amsfonts,bm}

\def\eqref#1{equation~\ref{#1}}
\def\1{\bm{1}}

\DeclareMathAlphabet{\mathsfit}{\encodingdefault}{\sfdefault}{m}{sl}
\SetMathAlphabet{\mathsfit}{bold}{\encodingdefault}{\sfdefault}{bx}{n}

\usepackage{amsmath,amssymb,booktabs,graphicx,multirow,xcolor,tabularx}
\usepackage[T1]{fontenc}
\usepackage{url,xspace}
\usepackage[hidelinks]{hyperref}
\hypersetup{
  pdftitle={Embodied Skill Bench: Benchmarking How Well Agents with Embodied Skills Perform on Physical Tasks},
  pdfauthor={Zhi Han, Zihan Guo, Chenxi Zeng, Zicai Cui, Yue Xin, Ming Zhou, Dongzhan Zhou, Weinan Zhang, Yang Li},
  pdfsubject={EmbodiedSkillBench preprint}
}
\ifdefined\pdfminorversion\pdfminorversion=7\fi
\ifdefined\XeTeXversion\fi

\newcommand{\ESB}{\textsc{EmbodiedSkillBench}\xspace}

\newcommand{\libero}{\textsc{LIBERO-Pro}}

\title{Embodied Skill Bench: Benchmarking How Well Agents with Embodied Skills Perform on Physical Tasks}

\author{%
  \fontsize{10}{12}\selectfont
  \textbf{Zhi Han}\textsuperscript{1,2,*}\quad
  \textbf{Zihan Guo}\textsuperscript{3,2,*}\quad
  \textbf{Chenxi Zeng}\textsuperscript{4,2}\\[3pt]
  \textbf{Zicai Cui}\textsuperscript{1,2}\quad
  \textbf{Yue Xin}\textsuperscript{5,2}\quad
  \textbf{Ming Zhou}\textsuperscript{2}\quad
  \textbf{Dongzhan Zhou}\textsuperscript{2}\\[3pt]
  \textbf{Weinan Zhang}\textsuperscript{1,2,\textdagger}\quad
  \textbf{Yang Li}\textsuperscript{1,2,\textdagger}\\[8pt]
  {\fontsize{9}{11}\selectfont
  \textsuperscript{1}Shanghai Jiao Tong University\\
  \textsuperscript{2}Shanghai Artificial Intelligence Laboratory\\
  \textsuperscript{3}Sun Yat-Sen University\quad
  \textsuperscript{4}Jilin University\\
  \textsuperscript{5}Beijing University of Aeronautics and Astronautics\\[5pt]
  \textsuperscript{*}Equal contribution.\quad
  \textsuperscript{\textdagger}Corresponding authors.}%
}
\date{}

\begin{document}
\maketitle

\begin{abstract}
LLM agents increasingly control robots using VLA policies, perception models, and procedural knowledge.
These modules are difficult to reuse because their interfaces, embodiment assumptions, and conditions of use differ, and their benefits on physical tasks remain unclear.
We package them as embodied skills that extend agent skills with explicit embodiment assumptions and conditions of use, building 250 action, perception, and experience skills from 1,052 resources.
On this basis we introduce \ESB, which runs an agent in a closed observation-action loop and records success, actions, tokens, and wall-clock time.
It supports three modes: native control, skill-only execution, and Agent-Directed Reuse (ADR), a mode we propose in which the agent keeps its native controls and decides for each skill whether to invoke, modify, or ignore it.
Across four agents on 100 perturbed LIBERO-Pro tasks, we find that skills help an agent only as far as they exceed what it can already do. 
Skill-only execution lifts GPT-5.6-Sol from 11\% to 63\% and Kimi-K3 and Claude-Opus-5 by about 20 points, while cutting their token usage by 58\% to 85\%.
Yet all four agents then end up between 61\% and 64\%, and GPT-6-Astra falls from 92\% under native control to 61\%.
This suggests that as agents grow stronger, the role of embodied skills shifts from supplying missing capability to reducing cost: GPT-6-Astra already solves most tasks through native control, so restricting it to skills removes more than it adds.
Under ADR, which lets it keep native control and call skills selectively, GPT-6-Astra recovers to 85\%, within 7 points of native control, while using 14.1\% fewer tokens and 11.9\% less time.
With Jev as a skill choosing motion directions and gripper actions, it reaches 81\% with 50\% fewer tokens and 57\% less time.
The code is available at \url{https://github.com/HanZhi306/EmbodiedSkill-Bench}.
\end{abstract}

\section{Introduction}
\label{sec:intro}
Large language model agents increasingly provide high-level robot control. Instead of mapping instructions directly to movements through a single policy, an agent can decompose tasks, call specialized modules, and revise plans after observing results. The language model handles task-level reasoning while robot-specific modules handle perception and movement. SayCan selects pretrained skills according to what is feasible in the scene~\citep{ahn2022saycan}, while Code as Policies generates programs that combine perception and control~\citep{liang2023codepolicies}. Recent harnesses connect agents to tools and environments, adding planning, memory, or recovery around frozen vision-language-action (VLA) policies~\citep{zhang2026harnessvla,ding2026zetta}. These policies translate images and language instructions into robot actions without being retrained by the agent.
Collectively, these developments have produced a growing ecosystem of reusable embodied resources, including native controllers, VLA models, visual perception tools, and practical guidance distilled from earlier robot runs. These resources provide knowledge that a general-purpose agent needs when working on physical tasks.%

\begin{figure}[t]
\centering
\includegraphics[width=\linewidth]{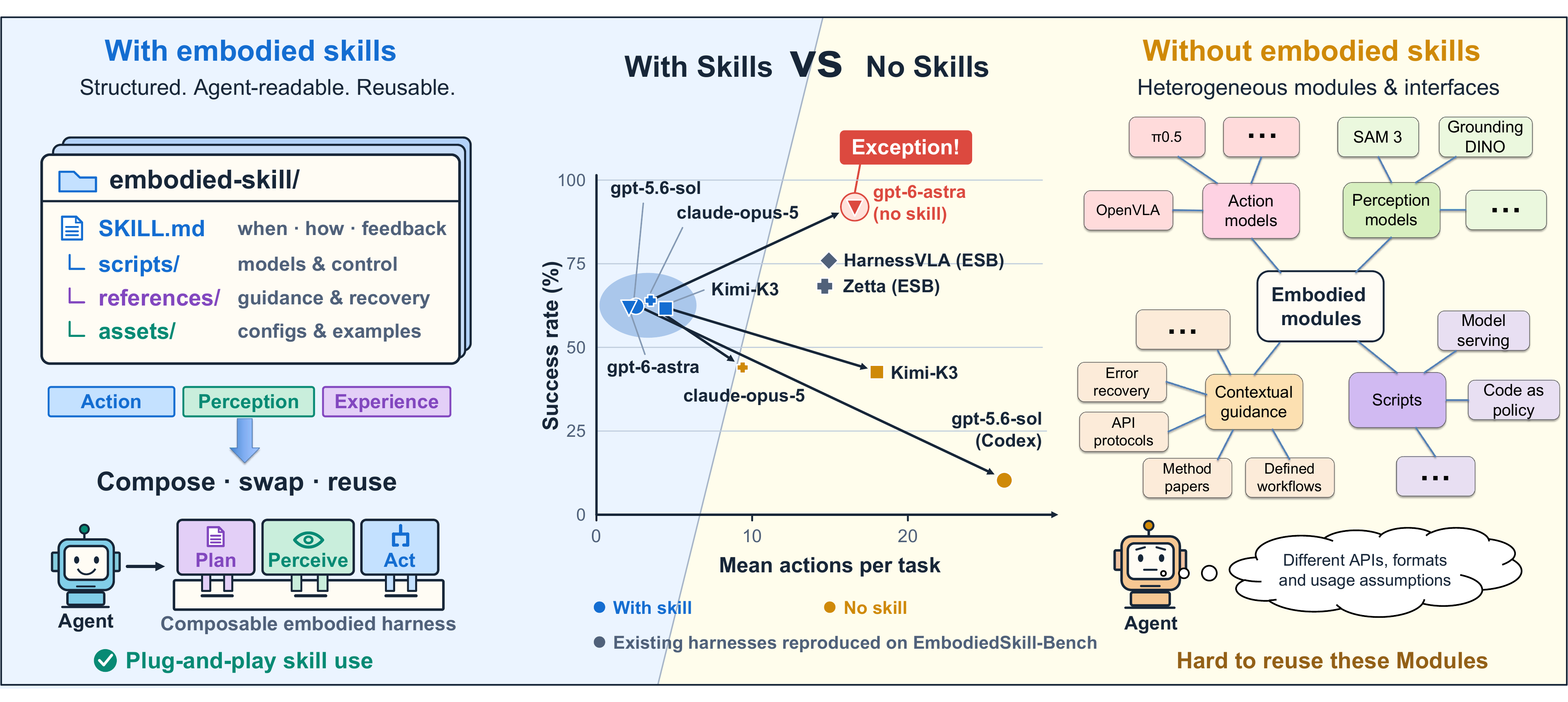}
\caption{\textbf{EmbodiedSkillBench: Standardizing embodied modules and evaluating skill use.} \textbf{Left:} Embodied skills package action, perception, and experience resources in a common, agent-readable format for composition and reuse. \textbf{Right:} Without this structure, agents must assemble workflows from heterogeneous modules and interfaces. \textbf{Center:} Arrows connect each agent's with-skill and no-skill results. Skills reduce mean action counts for all four agents and improve success for gpt-5.6-sol, Kimi-K3, and claude-opus-5; gpt-6-astra achieves higher success without skills.}
\label{fig:motivation}
\end{figure}

Recent work has explored reusable agent skills, embodied agent frameworks, and benchmarks for physical task execution. SkillsBench evaluates the utility of reusable agent skills, but covers only digital tasks, leaving the physical constraints and execution feedback of embodied skill use unexamined. ~\citep{li2026skillsbench} HarnessVLA and Zetta manually integrate a fixed set of modules into a method-specific harness, rather than making these modules independently reusable across agents and workflows.~\citep{zhang2026harnessvla,ding2026zetta} CaP-X and LIBERO-Pro evaluate agents or policies under specified control configurations, but configuration-level scores alone do not isolate how much these various and different modules contribute to an agent's performance.~\citep{fu2026capx, zhou2026liberopro}

As Figure 1 illustrates, the central challenge is that these heterogeneous embodied modules are hard for agents to reuse. Differences in APIs, data formats, embodiment assumptions, and execution conditions require an agent to determine how to invoke each module, when it applies, and how to interpret its feedback. Without a common way to expose these requirements, modules remain tied to individual integrations. Consequently, how effectively agents can reuse the available resources remains unclear. Answering this question requires a common representation that preserves module-specific physical constraints while supporting composition and reuse.

To address these gaps, we propose \ESB, a benchmark for evaluating how effectively and efficiently agents use reusable embodied modules to solve physical tasks. \ESB supports controlled comparisons across agents and skill configurations through a shared multi-turn interaction protocol, allowing agents to invoke modules and revise their plans based on physical feedback. To support this evaluation, we formulate embodied skills as a common representation that extends agent skills~\citep{anthropic2025skills,agentskills2025} with explicit embodiment assumptions, interface requirements, applicability conditions, and guidance for interpreting execution feedback. Using an autonomous agent workflow, we construct an Embodied Skill Repository of 250 skills from resources spanning four VLA model families with five model profiles, three perception models, eight programmed control interfaces, and procedural knowledge. We check the generated skills through scripts against the common protocol for skill component completeness and the availability of their declared model and control interfaces. The repository covers three complementary roles: action skills execute robot behaviors, perception skills interpret observations, and experience skills guide planning and recovery. By comparing agents with and without skills on the same tasks and linking skill-use traces to task success and resource consumption, \ESB measures the benefits and costs of embodied skill use.

We evaluate four agents on 100 LIBERO-Pro task cases spanning language, object, task, and swap perturbations, comparing performance with and without skills and examining different skill compositions. Embodied skills improve success by 19-52 points for GPT-5.6-Sol, Kimi-K3, and Claude-Opus-5 while reducing token usage by up to 85\%. Skill composition also matters: five configurations evaluated with GPT-5.6-Sol yield success rates ranging from 36\% to 77\%.  In particular, GPT-6-Astra achieves 92\% with native controls but only 61\% with skills, suggesting that many capabilities supplied by existing embodied modules have been trained and internalized in it. Motivated by this finding, we propose Agent-Directed Reuse, which allows agents to inspect, adapt, combine, or bypass skills while retaining native controls. This raises GPT-6-Astra's success to 85\%, recovering 24 percentage points over the with-skill setting while reducing tokens by 14.1\% and runtime by 11.9\% relative to native control setting, with a seven-point success tradeoff. We further integrate Jev as a waypoint-execution skill, allowing GPT-6-Astra to retain planning and recovery while delegating local motion decisions. This configuration achieves 81\% success and reduces mean runtime from 20.52 to 8.74 minutes relative to native control.

Our contributions are:
\begin{itemize}
    \item We formulate embodied skills as a common abstraction for heterogeneous embodied modules and establish an Embodied Skill Repository of 250 skills covering action, perception, and procedural guidance for physical tasks.
    \item We introduce \ESB, a framework for evaluating how well agents with embodied skills perform on physical tasks. It supports multiple agents and flexible skill composition within a multi-turn action-observation loop.
    \item Our experiments show that embodied skills improve task success and reduce the execution costs across different agents. A further study of GPT-6-astra motivates us to propose a novel interaction paradigm with skills: Agent-Directed Reuse, enhancing efficiency of agent on physical, with a measurable tradeoff in task success.
\end{itemize}

\section{Related Work}
\paragraph{Agents for physical control.} Language-model agents connect instructions to robot actions through skill selection, code generation, and feedback-driven planning~\citep{ahn2022saycan,liang2023codepolicies,huang2023innermonologue}. These approaches differ in the level of control delegated to the agent: some expose complete skills, while others require composing lower-level operations. Multimodal grounding further connects reasoning to physical observations and motion objectives~\citep{driess2023palme,huang2023voxposer}, while CaP-X examines how control abstractions affect coding agents~\citep{fu2026capx}. \ESB focuses on evaluating how well agents perform with these embodied modules on physical tasks.

\paragraph{Embodied skills and harnesses.} Generalist robot policies provide reusable language-conditioned control capabilities~\citep{brohan2023rt2,brohan2023rt1,ghosh2024octo,kim2025openvla}, but remain vulnerable to perturbations~\citep{zhou2026liberopro}. Agent harnesses and skill libraries add memory, recovery, and reusable procedures around robot controllers~\citep{zhang2026harnessvla,ding2026zetta,lu2026aspire,zhang2026rats}. However, these resources differ in their interfaces and usage assumptions, making their combination a separate challenge. \ESB standardizes heterogeneous capabilities as skill packages and compares how different agents use them, with skill provision and the surrounding harness specified explicitly.

\paragraph{Benchmarks for physical tasks.} RLBench, CALVIN, and LIBERO evaluate manipulation, long-horizon control, and knowledge transfer~\citep{james2020rlbench,mees2022calvin,liu2023libero}. LIBERO-Pro, VLABench, and EmbodiedBench further probe robustness, language understanding, and embodied reasoning~\citep{zhou2026liberopro,zhang2024vlabenchlargescalebenchmarklanguageconditioned,yang2025embodiedbenchcomprehensivebenchmarkingmultimodal}. These suites offer complementary tests of what a policy or agent can accomplish across tasks and changing conditions. SkillsBench and SkillRet evaluate skill utility and retrieval in digital tasks~\citep{li2026skillsbench,kang2026skillret}. Physical execution additionally requires compatible module interfaces and feedback that supports verification or recovery. Retrieval quality alone therefore does not establish that the selected capabilities can be used effectively. \ESB makes skill provision explicit in physical tasks, linking module use to task success and execution effort. This enables comparison of the benefits and costs of supplied capabilities under matched task conditions.

\section{Formulation}
\label{sec:formulation}
We define an embodied skill as a discoverable, progressively loadable, and reusable agent skill supporting interaction with a physical environment.

We represent a package as:
\begin{equation}
    s=\langle m,I,S,R,A\rangle
\end{equation}
where $m$ is metadata specifying the skill name and capability description; $I$ contains instructions for skill use, feedback interpretation, and recovery; $S$ contains scripts for control function and invocation of action and perception models; $R$ contains method descriptions, model-use documentation, and procedural experience; and $A$ contains assets such as model descriptors, configurations, interface schemas, and calibration data. Metadata and instructions are recorded in SKILL.md; the optional $S$, $R$, and $A$ components are organized in scripts, references, and assets, respectively, following the open Agent Skills format~\citep{agentskills2025}.

\paragraph{Embodied module types.} An embodied module is a source resource that contributes executable capability, perceptual evidence, or procedural knowledge for interaction with a physical environment. We distinguish three types by their resource form. \textbf{Model modules} include action and perception models. \textbf{Code modules} implement scripts for control function, observation processing, model invocation, and runtime adaptation. \textbf{Text modules} comprise context recording methods, interface documentation, procedural experience, and guidance for planning, verification, and recovery. This classification describes the source resources, while the operational role of an assembled skill depends on how those resources are combined and used.

\paragraph{Mapping modules to components.} $S$ contains Control and perception wrappers or interface declarations; $R$ method references, repository documentation, and experience material; $A$ model descriptors and supporting configurations. Model weights and serving runtimes remain external dependencies rather than being copied into each package. From these resources, the authoring agent derives $I$: applicability conditions, invocation steps, feedback interpretation, and bounded recovery or stopping rules. It then writes $m$ to summarize the skill's purpose and trigger conditions. For example, a grasping skill combines a control interface in $S$, the corresponding model profile and input schema in $A$, and model-use or recovery guidance in $R$; $I$ explains how to attempt and verify the grasp, while $m$ makes this capability discoverable. The mapping is many-to-many: one module can support multiple skills, and a skill can combine resources from several sources.

\begin{figure}[t]
\centering
\includegraphics[width=\linewidth]{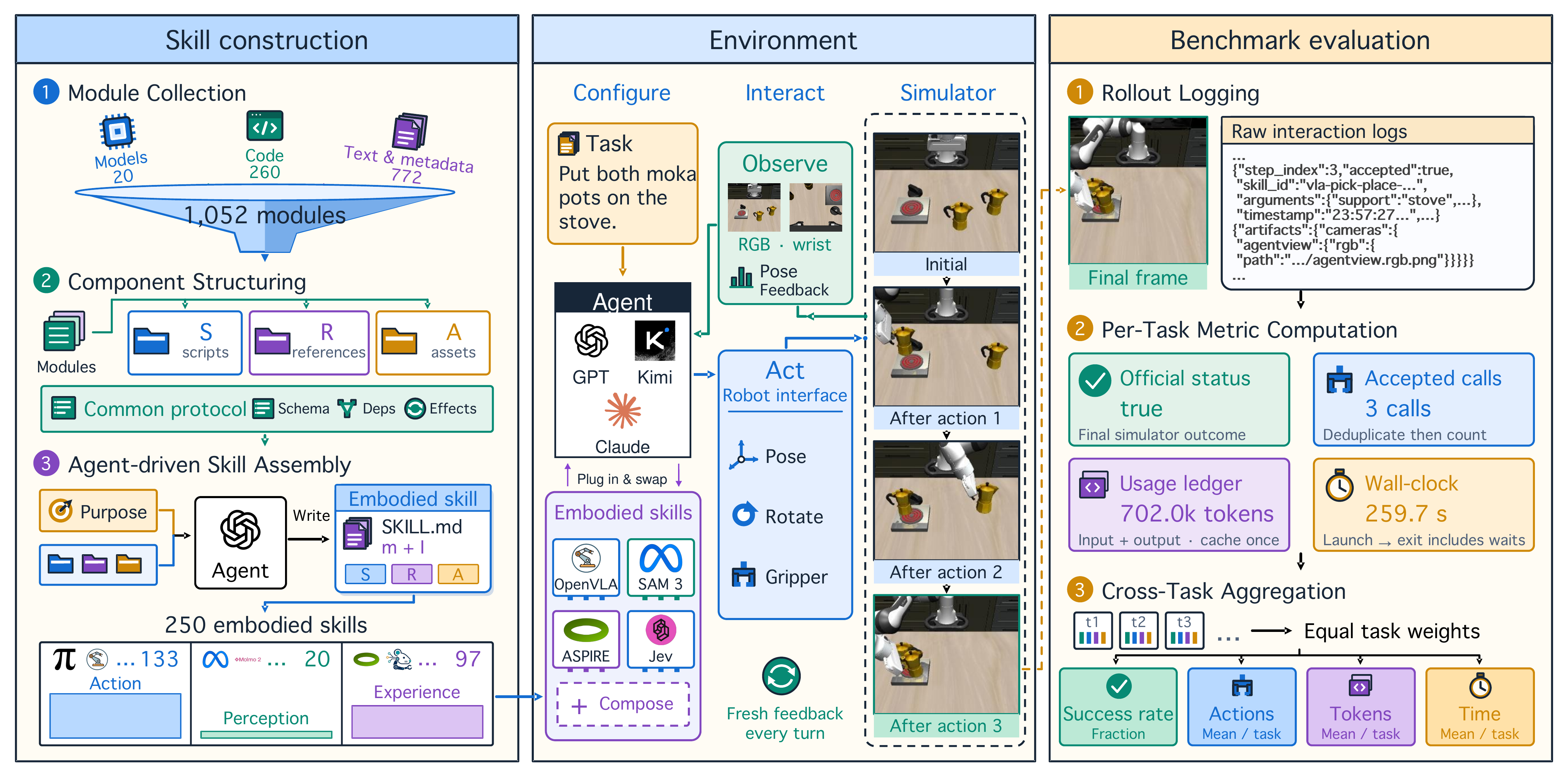}
\caption{\textbf{\ESB overview:} Skill construction collects heterogeneous modules, structures them under a common protocol, and uses an agent to assemble reusable embodied skills. The environment lets agents compose selected skills into task-specific harnesses and interact with a simulator through a multi-turn observation-action loop. Benchmark evaluation logs the resulting interactions, computes per-task success, actions, tokens, and time, and aggregates them across tasks.
}
\label{fig:framework}
\end{figure}

\section{EmbodiedSkillBench}
\label{sec:benchmark}
\ESB evaluates how effectively agents use embodied skills on physical tasks. It brings together the Embodied Skill Repository, a configurable agent-environment interface, and a trace-based evaluator. Figure~\ref{fig:framework} summarizes how embodied modules are assembled into skills, work in the embodied environment, and evaluated through multi-turn physical interaction.

The repository contains 250 embodied skill packages: 133 action skills (53.2\%) for VLA-based and programmed control, 20 perception skills (8.0\%) for localization, segmentation, pointing, and grasp proposals, and 97 experience skills (38.8\%) providing guidance for planning, verification, and recovery.

Benchmark runs combine an agent and a selected skill set with a physical task under a shared observation-action protocol. The current evaluation uses 100 \libero{} task cases spanning multiple manipulation domains and perturbations. Task success, accepted actions, recorded tokens, and elapsed time jointly characterize the effectiveness and execution effort of each configuration.

\subsection{Skill Construction}
\label{sec:roles}
The construction workflow in Figure~\ref{fig:framework} comprises module collection, component structuring, and agent-assisted skill assembly. We describe the source inventory and the concrete package scopes in the current Embodied Skill Repository.

\paragraph{Module collection.} The corpus contains 1,052 deduplicated resource records: 227 control resources, 33 perception-interface records, 20 model descriptors, 417 experience materials, 23 publication references, 13 repository references, and 319 supporting assets. We create a card for each record, containing its type, functional description, and source. The main repository's VLA interfaces cover four model families through five profiles: two $\pi_{0.5}$ checkpoints (RLinf and Madokalif), OpenVLA-OFT, SmolVLA, and X-VLA~\citep{physicalintelligence2025pi05,kim2025oft,shukor2025smolvla,zheng2025xvla}. Perception resources support SAM3, Molmo2, and Contact-GraspNet~\citep{carion2026sam3segmentconcepts,clark2026molmo2,sundermeyer2021contactgraspnet}. The 20 descriptor records include configuration and source variants. Experience sources include model-use instructions, manipulation and recovery procedures, and task-state verification guidance.

\paragraph{Component structuring.} Applying the mapping in Section~\ref{sec:formulation} groups the inventory into 260 script resources ($S$), 453 reference resources ($R$), and 339 assets ($A$). A common access protocol exposes resource reading and native-call preparation while retaining input schemas, dependencies, coordinate conventions, model profiles, and execution limits. For action-skill assembly, we retain eight task scopes---grasp, place-in, place-on, place-next-to, open, close, toggle, and push---across five VLA profiles and three budgets (base, brief, extended). Two concrete $\pi_{0.5}$ grasping interfaces and three general-subgoal interfaces also contribute to the VLA package. Six programmed interfaces for end-effector and gripper control, along with two mobile-base interfaces, provide the remaining action scopes. For perception, two general SAM3 interfaces and 16 queries specialized by target, view, resolution, and confidence are kept distinct, alongside one Molmo2 pointing interface and one Contact-GraspNet grasp-proposal interface. These perception skills use public observations without changing simulator state.

\paragraph{Agent-assisted skill assembly.} Given an operational purpose, an authoring agent selects compatible resources from the structured inventory and reads their interface contracts and supporting material. It writes instructions $I$ specifying applicability, invocation steps, feedback interpretation, verification, and bounded recovery or stopping rules, and metadata $m$ describing the capability and its trigger conditions. The assembly plan links these instructions to the selected $S$, $R$, and $A$ components. Materialization writes SKILL.md and the associated resource directories, records dependencies, and preserves source links. For skills providing procedural guidance, the instructions define read-only decision procedures and identify the action and perception capabilities required to carry them out; these dependencies must be selected separately.

\subsection{Environment}
\label{sec:environment}
As shown in the middle panel of Figure~\ref{fig:framework}, a run is configured with a task description, an agent, and a composable set of embodied skills. Through the shared observation interface (Observe), the agent reads scene and wrist RGB images, robot state, and execution feedback, then uses available skills or native robot controls to act through the robot interface (Act). Actions may specify end-effector motion, rotation, or gripper control, or invoke an available VLA skill. After each accepted action, the simulator returns fresh observations to support verification and replanning, forming a multi-turn loop until task termination or budget exhaustion.

\subsection{Evaluator}
\label{sec:evaluator}
Evaluation has three stages. Rollout Logging records multi-turn interactions, models' input and output histories and usage records, timestamps, and official simulator feedback. Per-Task Metric Computation derives task success, action counts, language-model tokens, and elapsed time from these records. Success follows the official final task status; actions count accepted state-changing invocations after deduplication; tokens sum recorded input and output usage, counting cached input once; elapsed time spans agent launch to exit, including model calls, tools, and waiting. Cross-Task Aggregation reports official task success rate and mean actions, tokens, and elapsed time over 100Clean on \libero and each metric uses its corresponding task-level records. Together, these measures capture task effectiveness and execution effort across configurations.

\section{Experiments and Analysis}
\label{sec:experiments}
Our experiments focus on three research questions:
\begin{description}
    \item[RQ1 (Effectiveness)] Do embodied skills improve agents' performance on physical tasks?
    \item[RQ2 (Composition)] How do different embodied skill sets selected from skill repository affect agents' performance on physical tasks?
    \item[RQ3 (Complementarity)] Can state-of-the-art models still benefit from embodied skills?
\end{description}

\subsection{Experimental setup}
\paragraph{Agents and harnesses.} We evaluate gpt-5.6-sol and gpt-6-astra using Codex CLI v0.148.0. We also evaluate Kimi-K3 through Kimi Code CLI v0.40.0 and claude-opus-5 through Claude Code v2.1.186. RQ1 compares gpt-5.6-sol, Kimi-K3, and claude-opus-5; RQ2 compares five skill configurations with gpt-5.6-sol; RQ3 separately studies gpt-6-astra.

\paragraph{Skill configurations.} In the \textbf{Selected} with-skill setting, gpt-5.6-sol has access to all 250 embodied skills and it is required to review the repository and select skills for each task individually. The resulting task-conditioned skill sets comprise action, perception, and experience skills. The gpt-5.6-sol, Kimi-K3, and claude-opus-5 settings receive the same selected skill set for each task, enabling comparison under common skill provision. For gpt-5.6-sol, we also evaluate \textbf{BM25} lexical retrieval~\citep{robertson2009bm25} and \textbf{Embedding} retrieval using BGE-small-en-v1.5. Both use the same 250 candidates and 481 public task-and-capability queries, taking the top-ranked skill per query and deduplicating within each task. They provide 4.23 and 3.82 skills per task on average, respectively. We additionally evaluate \textbf{HarnessVLA (ESB)} and \textbf{Zetta (ESB)}, our skill-based reconstructions of the original methods~\citep{zhang2026harnessvla,ding2026zetta}. These configurations compose their action modules and procedural guidance into fixed skill sets for gpt-5.6-sol. In the \textbf{no-skill} setting, agents have no embodied skills available and interact through \ESB observation and robot-control interfaces.

\paragraph{Tasks and Metrics.} We evaluate on Clean100, comprising 25 base tasks from \libero{} crossed with language, object, task, and swap perturbations~\citep{zhou2026liberopro}. The 100 task cases span LIBERO-10, Goal, Object, and Spatial (24, 28, 24, and 24 cases), covering multi-stage manipulation, varying goals, object identities, and spatial relations~\citep{liu2023libero}. Success rate is the primary metric; mean actions, tokens, and elapsed time measure execution effort(Section~\ref{sec:evaluator}).

\subsection{RQ1: Effectiveness of embodied skills}
\label{sec:rq1}

\begin{table}[t]
\caption{Skill effects on Clean100 (seed 26). L10, Goal, Object, and Spatial contain 24, 28, 24, and 24 tasks; Overall pools all 100. SR is success rate (\%); Act., Tok., and Time are mean agent-level actions, recorded language-model tokens (millions), and elapsed minutes, including failed tasks. Boldface marks the better value within each model pair (both for ties). Annotations below with-skill values show relative changes against no skill: green $\uparrow$ denotes SR gains, green $\downarrow$ reductions in Act., Tok., or Time, red $\downarrow$ SR declines, and gray 0.0\% no change. A dash (-) indicates an undefined relative change when the no-skill baseline is zero.}
\label{tab:main}
\centering
\fontsize{7}{8}\selectfont
\setlength{\tabcolsep}{1pt}
\renewcommand{\arraystretch}{1.15}
\newcommand{\esbup}[1]{\textcolor{green!45!black}{{\tiny$\uparrow$}#1}}
\newcommand{\esbsaving}[1]{\textcolor{green!45!black}{{\tiny$\downarrow$}#1}}
\newcommand{\esbdown}[1]{\textcolor{red!75!black}{{\tiny$\downarrow$}#1}}
\newcommand{\esbflat}[1]{\textcolor{black!55}{#1}}
\newcommand{\esbmetric}[2]{\shortstack[c]{#1\\[-0.4pt]{\fontsize{5.5}{6}\selectfont\strut#2}}}
\newsavebox{\esbtableonebox}
\sbox{\esbtableonebox}{%
\begin{tabular}{@{}l@{\hspace{3pt}}l*{20}{c}@{}}
\toprule
& & \multicolumn{4}{c}{L10} & \multicolumn{4}{c}{Goal} & \multicolumn{4}{c}{Object} & \multicolumn{4}{c}{Spatial} & \multicolumn{4}{c}{Overall}\\
\cmidrule(lr){3-6}\cmidrule(lr){7-10}\cmidrule(lr){11-14}\cmidrule(lr){15-18}\cmidrule(lr){19-22}
Model & Skill & SR & Act. & Tok. & Time & SR & Act. & Tok. & Time & SR & Act. & Tok. & Time & SR & Act. & Tok. & Time & SR & Act. & Tok. & Time\\
\midrule
gpt-5.6-sol & no & 0.0 & 30.04 & 6.004 & 11.11 & 14.3 & 21.07 & 4.710 & 10.82 & 4.2 & 28.00 & 5.692 & 12.89 & 25.0 & 25.88 & 5.115 & 13.43 & 11.0 & 26.04 & 5.354 & 12.01\\
 & \esbmetric{with}{} & \esbmetric{\textbf{41.7}}{\esbflat{-}} & \esbmetric{\textbf{4.83}}{\esbsaving{83.9\%}} & \esbmetric{\textbf{1.846}}{\esbsaving{69.3\%}} & \esbmetric{\textbf{8.79}}{\esbsaving{20.9\%}} & \esbmetric{\textbf{67.9}}{\esbup{375.0\%}} & \esbmetric{\textbf{1.61}}{\esbsaving{92.4\%}} & \esbmetric{\textbf{0.455}}{\esbsaving{90.3\%}} & \esbmetric{\textbf{2.92}}{\esbsaving{73.0\%}} & \esbmetric{\textbf{79.2}}{\esbup{1800.0\%}} & \esbmetric{\textbf{1.25}}{\esbsaving{95.5\%}} & \esbmetric{\textbf{0.442}}{\esbsaving{92.2\%}} & \esbmetric{\textbf{2.81}}{\esbsaving{78.2\%}} & \esbmetric{\textbf{62.5}}{\esbup{150.0\%}} & \esbmetric{\textbf{1.96}}{\esbsaving{92.4\%}} & \esbmetric{\textbf{0.528}}{\esbsaving{89.7\%}} & \esbmetric{\textbf{3.32}}{\esbsaving{75.2\%}} & \esbmetric{\textbf{63.0}}{\esbup{472.7\%}} & \esbmetric{\textbf{2.38}}{\esbsaving{90.9\%}} & \esbmetric{\textbf{0.803}}{\esbsaving{85.0\%}} & \esbmetric{\textbf{4.40}}{\esbsaving{63.4\%}}\\
\midrule
Kimi-K3 & no & 4.2 & 21.25 & 3.471 & 69.63 & 46.4 & 15.54 & 2.545 & 48.41 & \textbf{75.0} & 16.42 & 2.176 & 30.21 & 45.8 & 18.92 & 3.132 & 52.44 & 43.0 & 17.93 & 2.819 & 50.10\\
 & \esbmetric{with}{} & \esbmetric{\textbf{41.7}}{\esbup{900.0\%}} & \esbmetric{\textbf{5.21}}{\esbsaving{75.5\%}} & \esbmetric{\textbf{1.541}}{\esbsaving{55.6\%}} & \esbmetric{\textbf{19.30}}{\esbsaving{72.3\%}} & \esbmetric{\textbf{60.7}}{\esbup{30.8\%}} & \esbmetric{\textbf{5.68}}{\esbsaving{63.4\%}} & \esbmetric{\textbf{1.348}}{\esbsaving{47.0\%}} & \esbmetric{\textbf{17.20}}{\esbsaving{64.5\%}} & \esbmetric{\textbf{75.0}}{\esbflat{0.0\%}} & \esbmetric{\textbf{1.21}}{\esbsaving{92.6\%}} & \esbmetric{\textbf{0.335}}{\esbsaving{84.6\%}} & \esbmetric{\textbf{4.98}}{\esbsaving{83.5\%}} & \esbmetric{\textbf{70.8}}{\esbup{54.5\%}} & \esbmetric{\textbf{5.17}}{\esbsaving{72.7\%}} & \esbmetric{\textbf{1.490}}{\esbsaving{52.4\%}} & \esbmetric{\textbf{19.65}}{\esbsaving{62.5\%}} & \esbmetric{\textbf{62.0}}{\esbup{44.2\%}} & \esbmetric{\textbf{4.37}}{\esbsaving{75.6\%}} & \esbmetric{\textbf{1.185}}{\esbsaving{58.0\%}} & \esbmetric{\textbf{15.36}}{\esbsaving{69.3\%}}\\
\midrule
claude-opus-5 & no & 8.3 & 9.21 & 3.356 & 25.13 & 46.4 & 7.25 & 2.295 & 20.03 & 70.8 & 11.75 & 3.196 & 18.05 & 50.0 & 9.71 & 2.526 & 16.74 & 44.0 & 9.39 & 2.821 & 19.99\\
 & \esbmetric{with}{} & \esbmetric{\textbf{41.7}}{\esbup{400.0\%}} & \esbmetric{\textbf{5.29}}{\esbsaving{42.5\%}} & \esbmetric{\textbf{1.249}}{\esbsaving{62.8\%}} & \esbmetric{\textbf{12.46}}{\esbsaving{50.4\%}} & \esbmetric{\textbf{60.7}}{\esbup{30.8\%}} & \esbmetric{\textbf{4.25}}{\esbsaving{41.4\%}} & \esbmetric{\textbf{0.536}}{\esbsaving{76.6\%}} & \esbmetric{\textbf{6.54}}{\esbsaving{67.3\%}} & \esbmetric{\textbf{83.3}}{\esbup{17.6\%}} & \esbmetric{\textbf{1.21}}{\esbsaving{89.7\%}} & \esbmetric{\textbf{0.192}}{\esbsaving{94.0\%}} & \esbmetric{\textbf{2.54}}{\esbsaving{85.9\%}} & \esbmetric{\textbf{70.8}}{\esbup{41.7\%}} & \esbmetric{\textbf{3.04}}{\esbsaving{68.7\%}} & \esbmetric{\textbf{0.667}}{\esbsaving{73.6\%}} & \esbmetric{\textbf{7.71}}{\esbsaving{54.0\%}} & \esbmetric{\textbf{64.0}}{\esbup{45.5\%}} & \esbmetric{\textbf{3.48}}{\esbsaving{62.9\%}} & \esbmetric{\textbf{0.656}}{\esbsaving{76.7\%}} & \esbmetric{\textbf{7.28}}{\esbsaving{63.6\%}}\\
\bottomrule
\end{tabular}%
}
\typeout{ESB-TABLE1-NATURAL-WIDTH: \the\wd\esbtableonebox}
\resizebox{\linewidth}{!}{\usebox{\esbtableonebox}}
\end{table}

\begin{figure}[t]
\centering
\includegraphics[width=\linewidth]{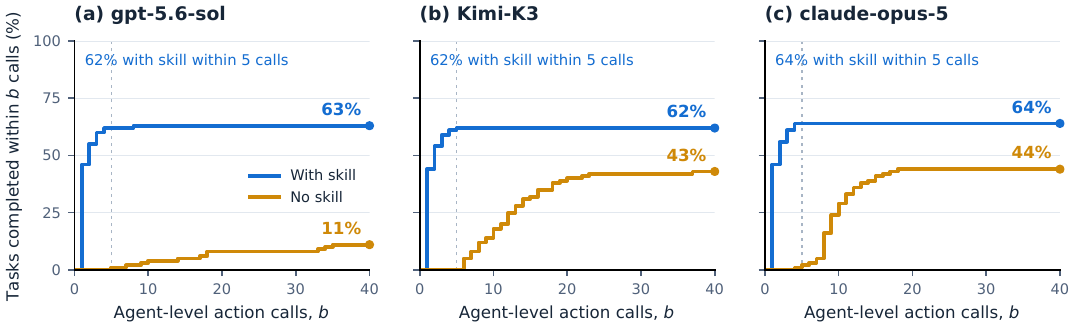}
\caption{Empirical completion profiles for (a) gpt-5.6-sol, (b) Kimi-K3, and (c) claude-opus-5. Each curve shows the percentage of all 100 tasks completed within $b$ agent-level action calls, on a linear horizontal axis. The dashed line marks five calls.}
\label{fig:rq1-actions}
\end{figure}

\paragraph{Task success and completion within few calls.} The three panels in Figure~\ref{fig:rq1-actions} show that embodied skills improve both final task success and completion within a small number of agent-level action calls. Embodied skills increase success by 52 points for gpt-5.6-sol, 19 for Kimi-K3, and 20 for claude-opus-5. Mean action calls per task decrease by 90.9\%, 75.6\%, and 62.9\%, respectively. Thus, the gains concern not only whether agents finish a task, but how much high-level interaction is needed to do. Their occurrence across three agents and multiple task domains supports embodied skills as reusable capabilities and guidance, rather than solutions restricted to one agent or task type.

\paragraph{Domain-specific strengths under shared skill provision.} As illustrated in Table~\ref{tab:main}, with the same selected skill sets, the three agents achieve similar overall success (62-64\%) but retain distinct domain strengths: gpt-5.6-sol leads on Goal, claude-opus-5 on Object, and Kimi-K3 and claude-opus-5 tie on Spatial. All three reach only 41.7\% on L10, indicating a shared challenge with multi-stage tasks. Common skill provision therefore does not eliminate model-specific differences in task interpretation and skill coordination.

\subsection{RQ2: Composition of embodied skills}
\label{sec:rq2}

\begin{table}[t]
\centering
\caption{Five skill configurations on Clean100 (seed 26), each evaluated on all 100 tasks. Success rates (SR, \%) are grouped by domain (24/28/24/24 tasks) and perturbation (25 tasks each). Actions, tokens, and time are per-task means, including failures. Boldface and underlining indicate the best and second best values in each column, respectively. All settings use gpt-5.6-sol with Codex; (ESB) denotes our skill-based reconstruction.}
\label{tab:rq2-selection}
\fontsize{8}{9}\selectfont
\setlength{\tabcolsep}{2pt}
\renewcommand{\arraystretch}{1.15}
\begin{tabular*}{\linewidth}{@{\extracolsep{\fill}}l*{12}{r}@{}}
\toprule
& \multicolumn{4}{c}{Domain SR (\%)} & \multicolumn{4}{c}{Perturbation SR (\%)} & \multicolumn{4}{c}{Overall}\\
\cmidrule(lr){2-5}\cmidrule(lr){6-9}\cmidrule(lr){10-13}
Setting & L10 & Goal & Object & Spatial & Lang. & Object & Task & Swap & SR (\%) & Actions & Tokens (k) & Time (min)\\
\midrule
Selected & \underline{41.7} & 67.9 & 79.2 & 62.5 & \textbf{92.0} & \textbf{84.0} & 40.0 & 36.0 & 63.0 & \textbf{2.38} & \textbf{803.2} & \textbf{4.40}\\
BM25 & 4.2 & 46.4 & 33.3 & 58.3 & 68.0 & 44.0 & 8.0 & 24.0 & 36.0 & \underline{8.09} & 2,731.6 & \underline{9.76}\\
Embedding & 25.0 & 39.3 & 45.8 & 58.3 & 72.0 & \underline{64.0} & 8.0 & 24.0 & 42.0 & 12.73 & 2,918.7 & 11.03\\
HarnessVLA (ESB) & \textbf{50.0} & \textbf{78.6} & \textbf{100.0} & \textbf{79.2} & \textbf{92.0} & 56.0 & \textbf{80.0} & \textbf{80.0} & \textbf{77.0} & 14.70 & \underline{2,113.8} & 14.12\\
Zetta (ESB) & 33.3 & \textbf{78.6} & \underline{91.7} & \underline{70.8} & 72.0 & 60.0 & \underline{68.0} & \underline{76.0} & \underline{69.0} & 14.51 & 2,399.7 & 16.66\\
\bottomrule
\end{tabular*}
\end{table}

\begin{figure}[t]
\centering
\includegraphics[width=\linewidth]{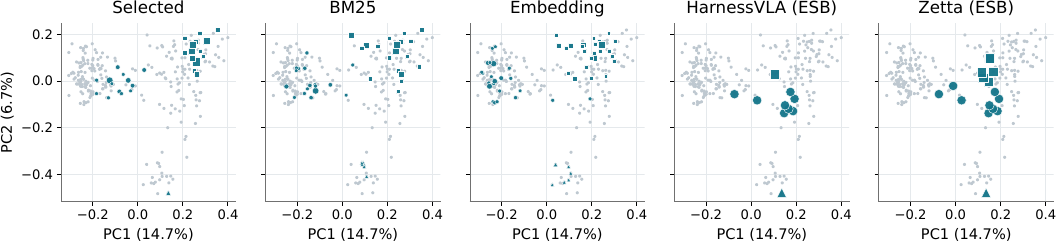}
\caption{Skill distributions under five provision settings. A shared PCA projects 384-dimensional BGE skill-document embeddings. Gray points show the reference skill documents; teal points show selected skills, with area increasing with the number of tasks selecting each skill. Circles, triangles, and squares denote action, perception, and experience skills. The panels contain 28, 41, 54, 10, and 17 distinct selected skills, respectively, across 100 tasks. The projection retains 21.5\% of embedding variance and illustrates document semantics rather than behavioral equivalence.}
\label{fig:rq2-selection}
\end{figure}

We evaluate five skill configurations with gpt-5.6-sol on the same 100 tasks: task-conditioned \textbf{Selected} sets, \textbf{BM25} lexical retrieval, \textbf{Embedding} retrieval, and fixed \textbf{HarnessVLA (ESB)} and \textbf{Zetta (ESB)} compositions. Embedding ranks candidates by cosine similarity between normalized BGE-small-en-v1.5 representations of public queries and skill documents. Figure~\ref{fig:rq2-selection} shows that Embedding selects 54 distinct skills across the task suite, compared with 41 for BM25 and 28 for Selected; the two reconstructed harnesses use fixed sets of ten and seventeen skills.

Table~\ref{tab:rq2-selection} shows that Embedding reaches 42\% success, compared with 36\% for BM25, 63\% for Selected, 77\% for HarnessVLA (ESB), and 69\% for Zetta (ESB). Its six-point advantage over BM25 accompanies more action calls (12.73 versus 8.09) and longer execution (11.03 versus 9.76 minutes). Broader retrieval across the repository therefore does not imply more efficient task completion. Embedding supplies no action skill on seven tasks, compared with twenty for BM25; all seven end without a physical action or task success. All they have access are native cotrollers. The wine-bottle case in Appendix~\ref{app:rq2-case} illustrates this distinction between relevant guidance and executable capability.

The domain results also rule out a uniform ranking of the retrieval methods. Embedding exceeds BM25 on L10 (25.0\% versus 4.2\%) and Object (45.8\% versus 33.3\%), falls below it on Goal (39.3\% versus 46.4\%), and matches it on Spatial (58.3\%). Across perturbations, it scores higher on language and object changes but matches BM25's 8\% on task changes and 24\% on swap. In contrast, HarnessVLA (ESB) reaches 80\% on both task changes and swap, while Selected leads on object perturbations at 84\%. These profiles motivate selection strategies that account for executable coverage and useful combinations of capabilities, alongside semantic relevance.

\subsection{RQ3: Complementarity with state-of-the-art models}
\label{sec:rq3}
\paragraph{Internalized competence and execution cost.} Following RQ1's gains across three agents, we examine complementarity with GPT-6-astra. Table~\ref{tab:rq3} shows 92\% success with no skill versus 61\% with skills, with native control outperforming skill-only execution in every domain. This gap suggests that GPT-6-astra may already have internalized some knowledge and capabilities represented by embodied modules, enabling task execution through native robot tools. Nevertheless, skills reduce mean actions by 86.50\%, recorded tokens by 75.51\%, and time by 83.43\% per task, revealing a tradeoff between success and execution cost.

\begin{table}[t]
\centering
\small
\caption{GPT-6-astra settings on Clean100 (seed 26). Boldface and underlining indicate the best and second best values in each column, respectively. Annotations next
to Agent-Directed Reuse values show relative changes from no skill setting:
red $\downarrow$ denotes lower SR, and green $\downarrow$ denotes
reduced actions, tokens, or time.}
\label{tab:rq3}

\setlength{\tabcolsep}{3pt}
\renewcommand{\arraystretch}{1.15}

\newcommand{\rqthreedown}[2]{%
  \rlap{\hspace{2pt}%
    {\scriptsize\textcolor{#1}{$\downarrow$\,#2}}}%
}

\resizebox{\linewidth}{!}{%
\begin{tabular}{
  @{}l
  *{4}{c}
  @{\hspace{6mm}}
  *{4}{>{\centering\arraybackslash}m{20mm}}
  @{\hspace{2.5mm}}
}
\toprule
& \multicolumn{4}{c}{Domain SR (\%)}
& \multicolumn{4}{c}{Overall} \\
\cmidrule(lr){2-5}
\cmidrule(lr){6-9}

Configuration
& L10
& Goal
& Object
& Spatial
& SR (\%)
& Actions
& Tokens (k)
& Time (min) \\

\midrule
With skill
& 37.5
& 67.9
& 79.2
& 58.3
& 61.0
& \textbf{2.24}
& \textbf{563.5}
& \textbf{3.40} \\

With Jev skill
& 58.3
& \underline{92.9}
& 79.2
& \underline{91.7}
& 81.0
& 20.90
& \underline{1,142.0}
& \underline{8.74} \\

No skill
& \textbf{70.8}
& \textbf{100.0}
& \textbf{100.0}
& \textbf{95.8}
& \textbf{92.0}
& 16.59
& 2,300.6
& 20.52 \\

\midrule
Agent-Directed Reuse
& \underline{62.5}
& 89.3
& \underline{95.8}
& \underline{91.7}
& \underline{85.0}\rqthreedown{red!75!black}{7.6\%}
& \underline{12.87}\rqthreedown{green!45!black}{22.4\%}
& 1,976.3\rqthreedown{green!45!black}{14.1\%}
& 18.09\rqthreedown{green!45!black}{11.9\%} \\

\bottomrule
\end{tabular}%
}
\end{table}

\paragraph{Agent-Directed Reuse.} This contrast motivates a new interaction paradigm: can an agent preserve its task-solving ability while using embodied skills to reduce execution effort? In \textbf{Agent-Directed Reuse}, GPT-6-astra may independently inspect, use, modify, combine, or ignore skills, avoiding prescribed workflows that constrain powerful models. This configuration reaches 85\% success, recovering 24 percentage points over the with-skill setting while remaining seven points below no skill. Relative to no skill, mean actions, recorded tokens, and elapsed time decrease by 22.4\%, 14.1\%, and 11.9\%, respectively. These results support agent-directed interaction as a promising route to complementarity.

\paragraph{Jev as an embodied skill.} Jev's specialized decision interface selects among specified options and returns their probabilities~\citep{typesafe2026jev}. Following EmbodiedJev's robot-control workflow~\citep{fbddcz2026embodiedjev}, we package Jev as a single waypoint-execution skill and benchmark it with GPT-6-astra. It selects discrete motion directions and gripper commands for agent-specified subgoals; GPT-6-astra retains visual grounding, task planning, and recovery. Success reaches 81\%, 20 percentage points above the original with-skill setting but below Agent-Directed Reuse and no skill. Mean runtime falls to 8.74 minutes despite more agent-level action calls. Success is high on Goal and Spatial but lower on L10; multi-stage tasks remain challenging.

\section{Conclusion}
\ESB benchmarks how well agents use embodied modules on physical tasks. Its common skill representation and reusable repository of action, perception, and experience resources support cross-agent comparisons and flexible skill selection under a shared interaction protocol.

Embodied skills can improve task success and reduce execution effort when they complement existing agent capabilities. Semantic relevance alone does not ensure executable, complementary skills; prescribed workflows may constrain agents capable of effective native control, such as GPT-6-astra. Agent-Directed Reuse and Jev integration on GPT-6-astra demonstrate efficiency gains from flexible reuse and selective delegation, although preserving native-control success remains challenging.

Future skill selection strategies should jointly consider task requirements, agent capabilities, and skill compatibility. Physical feedback could guide when agents delegate to a skill, resume direct control, or revise procedures. Retaining verified revisions across tasks could support continual adaptation, making embodied skills more responsive to changing tasks and agent capabilities.

\bibliography{references}
\bibliographystyle{esb_preprint}

\appendix
\section{RQ1: No-skill interface and a multi-object case study}
\label{app:rq1}

The following three case studies connect configuration choices to the actions and observations of individual tasks. Each comparison fixes the task ID and environment seed (26), and follows the evaluated configuration's own recorded rollout. Cases are selected for their explanatory trajectories: multi-object progress for RQ1, grounding and phase composition for RQ2, and controller switching and waypoint revision for RQ3. They illustrate mechanisms behind the aggregate results, rather than estimating their frequency. Outcomes come from the official task status. Action indices refer to state-changing agent calls. Tables report cumulative input plus output tokens, counting cached input once, and recorded elapsed time.

\subsection{The no-skill framework}
\label{app:noskill}

\paragraph{What the agent receives.}
The agent controls one persistent LIBERO-Pro episode through a typed, task-agnostic robot interface. Its initial message contains the instruction and the external-camera and wrist-camera RGB observations. Subsequent observations contain the current robot state, execution feedback, and official task status. The controller reference specifies coordinate frames, units, gripper semantics, and geometric tolerances. Task decomposition, target identification, grasp selection, and recovery decisions are made by the language model from these observations.

The native interface uses a Franka Panda with a joint-position controller. Cartesian commands pass through a frozen PyRoKi inverse-kinematics (IK) gate; invalid solutions are rejected before execution. Positions and translations use metres in the LIBERO world frame, with $+z$ pointing upward. Full orientations use quaternions ordered $(x,y,z,w)$. Camera calibration and pixel-depth queries provide geometric grounding of agent-selected pixels. They do not assign semantic identities to those pixels. The native RGB-D resolution is $256\times256$; the interface's higher-resolution view is a deterministic enlargement.

\begin{table}[htbp]
\centering\small
\caption{No-skill tool surface. Read-only calls leave the physical state unchanged. Each accepted state-changing call returns a fresh observation.}
\label{tab:no-skill-interface}
\begin{tabularx}{\linewidth}{@{}p{.39\linewidth}X@{}}
\toprule
Interface & Function supplied to the agent\\
\midrule
observation, end-effector state & Current instruction, RGB, status, end-effector pose, and gripper state.\\
camera calibration, pixel depth, pixel-to-world projection & Camera intrinsics/extrinsics, metric depth, and calibrated back-projection of a selected pixel.\\
gripper contact & Identity-free gripper-width/contact evidence; the agent checks the images to assess a grasp.\\
IK pose check & Read-only feasibility check of a proposed full pose.\\
relative end-effector motion, absolute end-effector motion, full-pose IK motion & Bounded relative translation, absolute Cartesian motion, and full-pose IK motion.\\
gripper control & Open or close the gripper while holding the measured arm joints.\\
\bottomrule
\end{tabularx}
\end{table}

\paragraph{Interaction and budgets.}
Read-only queries may precede an action. The first physical call in a model response ends that round, and the agent consumes its returned RGB before choosing another action in a later response. Task interaction in the no-skill condition is through the registered native tools. The gpt-5.6-sol, Kimi-K3, and claude-opus-5 cases below use Codex CLI, Kimi Code CLI, and Claude Code, respectively.

The following card presents excerpts from the staged no-skill system instructions. Throughout the appendix, interface and package identifiers are displayed as readable labels; prompt wording is otherwise retained.
\begin{center}
\includegraphics[width=\linewidth]{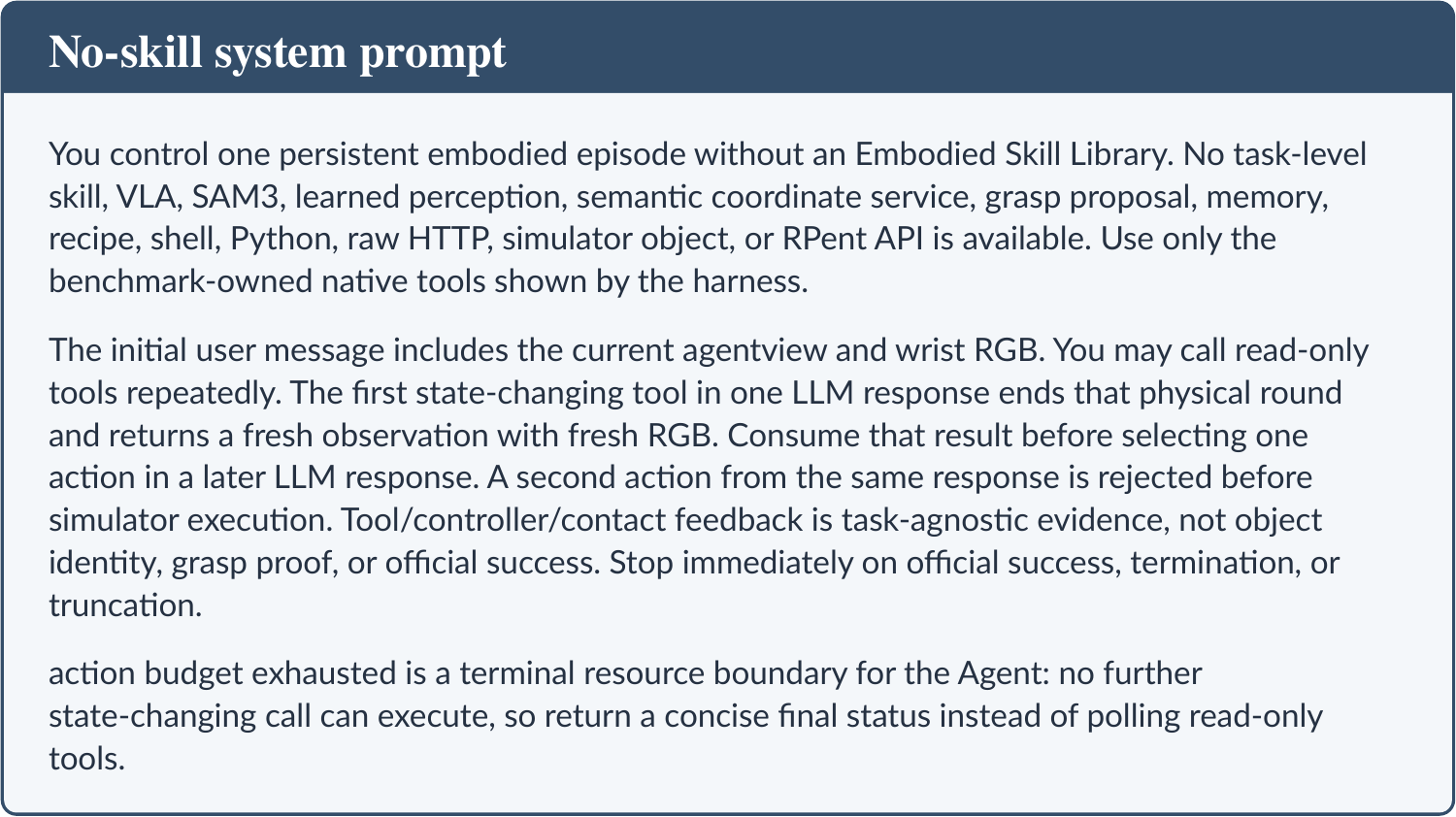}
\end{center}

\subsection{Case: placing both moka pots on the stove}
\label{app:rq1-case}

\paragraph{Task and supplied skills.}
The instruction is ``put both moka pots on the stove'' (LIBERO-10, object perturbation, task 08). Two yellow moka pots begin on the counter, to the right of the stove. Completion requires moving both objects, so a successful first transfer is only an intermediate state. The selected gpt-5.6-sol, Kimi-K3, and claude-opus-5 configurations receive the same four packages. Planning is supplied by esbench-decompose-atomic, with progress tracking in multi-object progress tracking. The action package vla-pick-place-on-openvla-oft-v1 executes a bounded OpenVLA-OFT transfer. The final experience package, support-placement verification, guides placement verification.

\begin{table}[htbp]
\centering\small
\caption{RQ1 case results on the same two-pot task, seed 26. Time is in minutes; tokens are in thousands.}
\label{tab:rq1-case}
\begin{tabular*}{\linewidth}{@{\extracolsep{\fill}}llrrr@{}}
\toprule
Configuration & Outcome & Actions & Tokens (k) & Time (min)\\
\midrule
gpt-5.6-sol / no skill & Failure & 40 & 7,250.6 & 13.31\\
gpt-5.6-sol / with skill & Success & 3 & 702.0 & 4.33\\
Kimi-K3 / no skill & Failure & 25 & 4,736.5 & 87.70\\
Kimi-K3 / with skill & Success & 3 & 317.5 & 3.35\\
claude-opus-5 / no skill & Failure & 13 & 3,036.4 & 19.46\\
claude-opus-5 / with skill & Success & 3 & 364.8 & 4.86\\
\bottomrule
\end{tabular*}
\end{table}

\paragraph{gpt-5.6-sol: a reusable transfer closes the execution gap.}
Without skills, the agent repeatedly approaches and closes around the pots, then moves toward the stove. The images show that these movements do not establish a retained payload: both pots remain on the counter at action 40, when the action budget ends. With skills, the agent first calls the pick-and-place package with object ``yellow moka pot nearest the stove'' and support ``stove.'' After this 160-step operation, the first pot is above the stove and official success remains false. The next call targets ``yellow moka pot remaining on the counter.'' Its returned image shows the first pot on the stove and the second held above the counter. A third call continues this remaining-object transfer, reaching official success after 64 further simulator steps. This is a three-call, 384-step rollout: progress tracking distinguishes the completed object from the remaining one, while the skill executes each contact-rich transfer.

\paragraph{Kimi-K3: recovery attempts versus retained multi-object progress.}
The no-skill rollout topples the first pot by action 5. It then attempts the second pot, which is also lying on the counter by action 18. At the configured 25-action boundary, neither pot is on the stove. In the skill-enabled rollout, Kimi first names the pot ``closer to the stove,'' then names the pot ``on the table'' in the two subsequent calls. The second transfer finishes on action 3, after 400 simulator steps. The two successful skill-enabled agents therefore follow the same observable structure: target one object, inspect the result, target the remaining object, and continue until official completion.

\paragraph{claude-opus-5: reusing the same transfer across two objects.}
Without skills, claude-opus-5 makes 13 calls involving end-effector motion and gripper control, but the rollout ends without completing the two-pot task. With the same four packages as the other agents, it first requests the pot ``closest to the stove,'' then the pot ``farthest from the stove,'' and finally the pot ``still on the wooden table.'' The three pick-and-place calls execute 160, 160, and 56 simulator steps, respectively, reaching official success after 376 steps. The with-skill rollout takes 4.86 minutes, compared with 19.46 minutes for the unsuccessful no-skill rollout. Here the same control package is reused with revised object descriptions as the scene changes.

\paragraph{Relation to RQ1.}
This case illustrates successful skill-enabled execution across all three RQ1 agents: each reuses a language-conditioned transfer while tracking which object remains to be moved. All three finish in three calls with skills, whereas their no-skill rollouts fail on this task. The shared packages therefore support both control execution and multi-object progress tracking. The target-grounding failure in Appendix~\ref{app:rq3-case} examines the complementary situation in which repeated delegation is insufficient. Neither case alone establishes the frequency of these behaviors across tasks.

\begin{figure}[p]
\centering
\includegraphics[width=\linewidth]{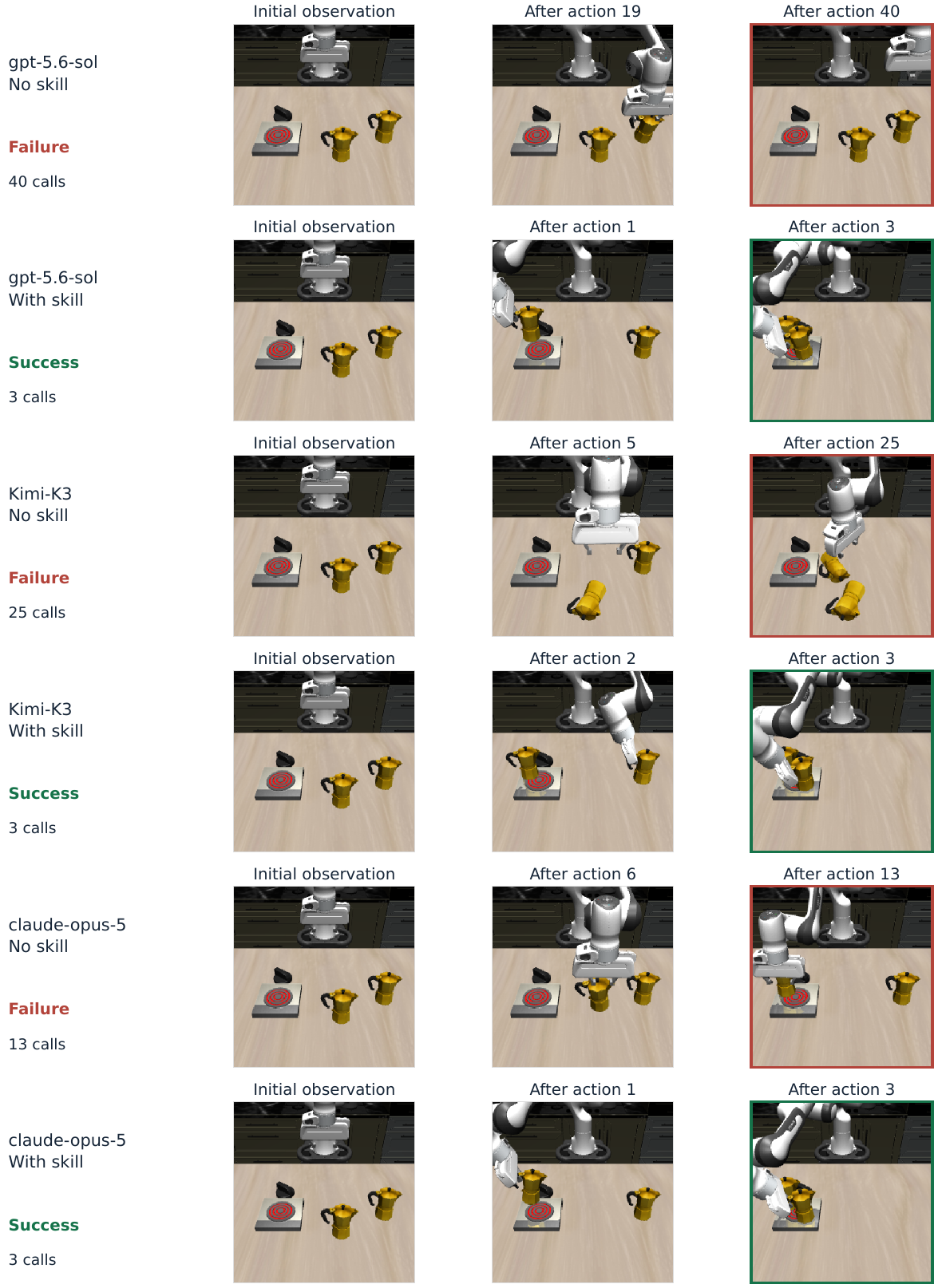}
\caption{\textbf{RQ1: both moka pots must reach the stove.} Each row follows one recorded rollout through its initial observation, a selected intermediate observation, and its final observation. gpt-5.6-sol, Kimi-K3, and claude-opus-5 fail without skills but complete both transfers in three calls with the same supplied packages. Action numbers identify the actual recorded calls; columns are not synchronized in physical time. Borders and outcome labels mark official success or failure.}
\label{fig:rq1-case}
\end{figure}

\clearpage
\section{RQ2: Skill selection and embodied harness composition}
\label{app:rq2}

\subsection{HarnessVLA (ESB) and Zetta (ESB)}
\label{app:harness-composition}

\paragraph{Common execution substrate.}
HarnessVLA (ESB) and Zetta (ESB) reconstruct the original control workflows~\citep{zhang2026harnessvla,ding2026zetta} as compositions of action, perception, and experience packages. Each package has an agent-readable skill instruction file; executable packages additionally provide a local execution contract. The agent reads a selected package, invokes it through the workspace launcher, and receives a new observation. An experience package supplies guidance for choosing and sequencing these operations. Both reconstructions install a fixed package set across tasks and use gpt-5.6-sol with a 40-action limit.

\begin{table}[htbp]
\centering\footnotesize
\caption{Mapping harness components to embodied skills. HarnessVLA (ESB) installs ten packages; Zetta (ESB) installs seventeen. Experience packages are read-only.}
\label{tab:harness-composition}
\begin{tabularx}{\linewidth}{@{}p{.24\linewidth}X@{}}
\toprule
Component & Embodied skill implementation\\
\midrule
Free-space motion (both) & move to, move pose, rotate wrist, rotate pitch.\\
Gripper transitions (both) & set gripper, release.\\
Local VLA contact (both) & pi0 pick, pi0 doubled; Zetta (ESB) additionally installs VLA object grasp.\\
Current-scene perception (both) & SAM3 object localization, invoked as a separate read-only query.\\
HarnessVLA (ESB) planning & HarnessVLA (ESB) memory-guided workflow, with a global rule reference and task-memory schema.\\
Zetta (ESB) online orchestration & Zetta (ESB) closed-loop evolution workflow, Zetta (ESB) runtime critic, Zetta (ESB) Role1 recovery review.\\
Zetta (ESB) campaign references & failure clustering, causal diagnosis, critic recovery proposal, gate protocol.\\
\bottomrule
\end{tabularx}
\end{table}

\paragraph{HarnessVLA (ESB): strategy retrieval, live grounding, and local retry.}
The memory-guided workflow packages abstract subgoal ordering, primitive operating ranges, and failure/repair rules. Its fixed ESB references specify, for example, free-space staging before a VLA contact call, checking gripper and image evidence before transport, and accounting for the offset between a grasped object and the end effector. Object identities, image locations, and geometric targets are bound to the current observation. The agent uses analytic skills for approach, lift, transport, descent, and retreat; the frozen VLA handles a bounded local contact phase. After a failure, the workflow directs the agent to inspect the last primitive, change a relevant factor, and retry that phase while preserving completed progress. The original method's VLA action abstraction is represented by the concrete pi0 pick and pi0 doubled packages.

\paragraph{Zetta (ESB): critic proposals under agent authority.}
The Zetta (ESB) composition adds a read-only temporal critic and a recovery-review role to the action substrate. The critic contract specifies a feature, comparison operator, threshold, dwell count, cooldown, and activation conditions. A fired critic proposes a bounded recovery. The agent, acting as Role1, checks the proposal and selects the action skills that execute an accepted recovery. The fixed ESB rule reference includes retention-loss and stalled-progress patterns, with re-grounding/re-grasping and local re-engagement procedures. Current observations determine their applicability; absolute object poses are not supplied as a motion oracle. The four campaign-role packages specify clustering, diagnosis, proposal, and gating contracts. The evaluated episodes use the fixed online composition; these campaign roles remain references during execution. This separates the packaged runtime workflow from an offline process for revising its rule set.

The following two excerpts state the operational division of labour and proposal authority in the staged system instructions; method names are displayed with the ESB suffix:
\begin{center}
\includegraphics[width=\linewidth]{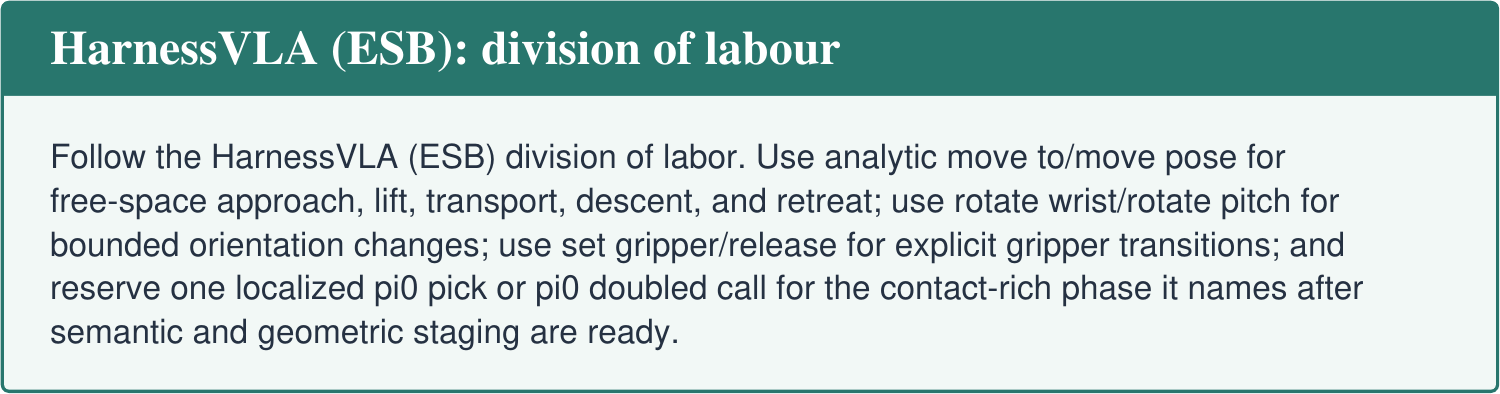}
\end{center}
\begin{center}
\includegraphics[width=\linewidth]{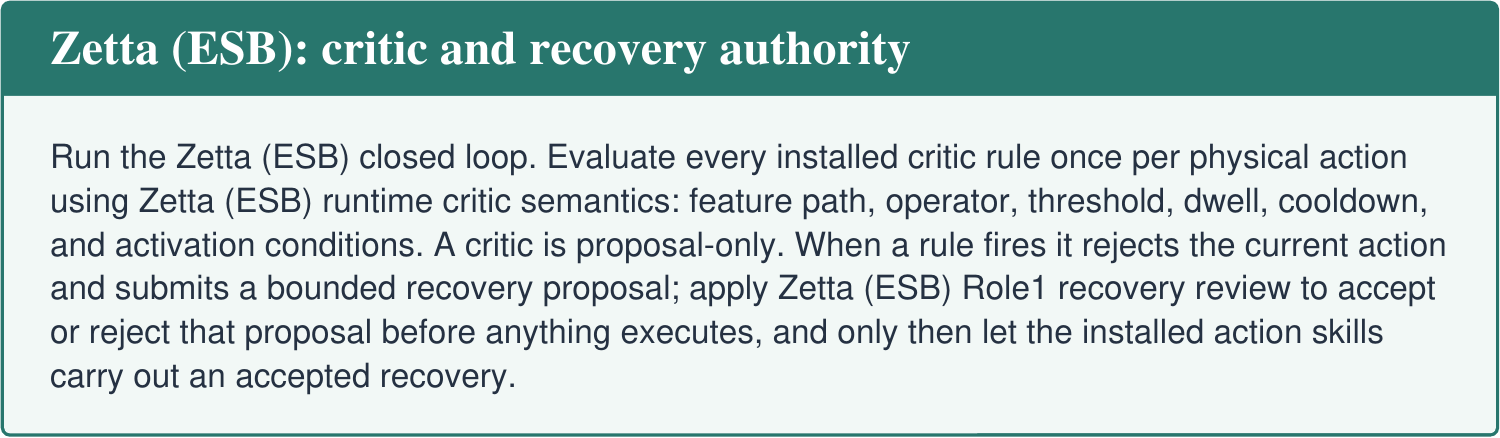}
\end{center}

\subsection{Case: placing a wine bottle on the drawer top}
\label{app:rq2-case}

\paragraph{Task and comparison.}
The instruction is ``Put the wine bottle on the top of the drawer'' (LIBERO-Goal, swap perturbation, task 02). A bottle and a bowl are visible beside the cabinet; its dark top surface must be distinguished from the slatted drawer face. Table~\ref{tab:rq2-case} compares all five RQ2 configurations on this task. In the Selected setting, gpt-5.6-sol receives three skills: esbench-decompose-atomic, vla-pick-place-on-openvla-oft-v1, and support-placement verification. BM25 retrieves X-VLA opening, VLA placement, and the same support-verification experience. Embedding retrieves three experience skills: narrow-container alignment, rim-collision recovery, and support-placement verification. The two embodied harnesses receive the fixed compositions in Table~\ref{tab:harness-composition}.

\begin{table}[htbp]
\centering\small
\caption{RQ2 case results on the wine-bottle task, seed 26. All rows use gpt-5.6-sol.}
\label{tab:rq2-case}
\begin{tabular*}{\linewidth}{@{\extracolsep{\fill}}llrrr@{}}
\toprule
Configuration & Outcome & Actions & Tokens (k) & Time (min)\\
\midrule
Selected & Failure & 2 & 592.8 & 3.75\\
BM25 & Failure & 16 & 3,989.0 & 15.42\\
Embedding & Failure & 0 & 72.1 & 0.56\\
HarnessVLA (ESB) & Success & 8 & 839.6 & 5.99\\
Zetta (ESB) & Success & 9 & 1,334.9 & 7.01\\
\bottomrule
\end{tabular*}
\end{table}

\paragraph{Selected: correct language does not ensure correct contact.}
The agent invokes the OpenVLA-OFT package twice with the wine bottle as the object and the drawer top as the support, using 20 chunks per call. In the returned images, the policy manipulates the neighbouring bowl while the bottle remains upright on the table. The rollout does not reach official success, and the agent's final report explicitly identifies the wrong-object grasp. This trace distinguishes an agent issuing the correct object name from a delegated policy physically acting on that object. Repeating the high-level request does not resolve this mismatch in the rollout.

\paragraph{BM25: an executable placement skill can still lack a useful recovery path.}
The retrieved placement package does grasp the bottle. However, it carries the bottle toward the slatted cabinet face instead of establishing it on the dark top. The agent repeatedly revises the destination language: calls 1--6 use ``top of drawer,'' calls 7--8 use ``drawer,'' and later calls specify the ``flat black top surface'' or ``black tabletop of drawer unit.'' All sixteen physical calls invoke VLA placement; the retrieved X-VLA opening skill is not used. The final image still shows the bottle at the slatted face and official success remains false. This case therefore exposes a utilization problem beyond selecting any executable action: the available and chosen operations must support the geometric correction demanded by the observed failure.

\paragraph{Embedding: relevant guidance without an executable action.}
The three retrieved packages provide read-only guidance but no action interface. The agent inspects the observation and identifies that no robot-changing skill is installed. It makes no physical action calls and reports failure, consistent with the official outcome. The run ends after 0.56 minutes with 72.1k tokens; the final observation is the unchanged initial scene. This case illustrates a failure of executable coverage before manipulation begins, distinct from the repeated unsuccessful placement calls in BM25. A short rollout here reflects an inability to act, not efficient task completion.

\paragraph{HarnessVLA (ESB): ground, stage, and complete a local grasp.}
The agent first issues two SAM3 queries, one for the wine bottle and one for the drawer's top surface. Action 1 moves the open gripper above the bottle. Action 2 delegates only ``grasp and lift the wine bottle directly below the gripper'' to pi0 pick, for twelve chunks. Its feedback reports an unmet kinematic postcondition, with the gripper still about 0.080\,m open. The agent closes the gripper explicitly at action 3 and lifts at actions 4--5; the images then show the bottle moving with the gripper. Actions 6--7 transport it above the cabinet and action 8 lowers it onto the top, where official success triggers. The frozen VLA supplies one contact phase, while the agent's selected motion and gripper skills finish the manipulation.

\paragraph{Zetta (ESB): feedback-guided phase composition.}
Zetta (ESB) also grounds the bottle and cabinet using two SAM3 queries, stages above the bottle, and makes one twelve-chunk pi0 pick call. After the same unmet grasp postcondition, it explicitly closes the gripper, lifts, transports with clearance, and lowers. Its ninth action is release, which triggers official success. This successful trace demonstrates the installed workflow's use of local feedback and explicit phase transitions. The observed correction is the gripper-and-lift continuation following the contact call; the outcome itself does not establish that a temporal critic intervention caused the success.

\begin{figure}[p]
\centering
\includegraphics[width=\linewidth]{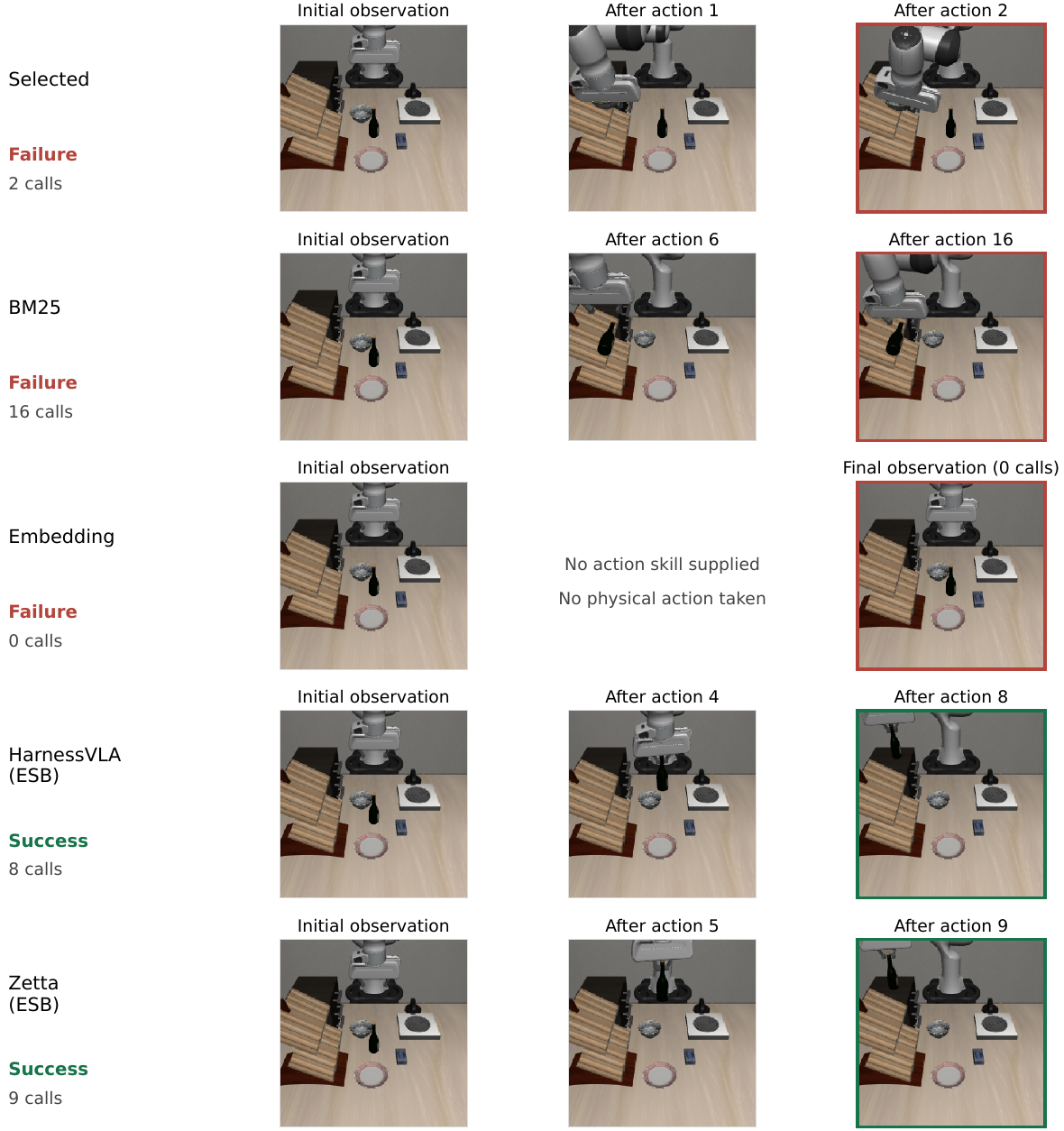}
\caption{\textbf{RQ2: wine bottle onto the drawer top.} The Selected agent leaves the bottle on the table while manipulating the bowl. BM25 moves the bottle to the slatted face. Embedding receives only experience skills and takes no physical action; its initial and final panels show the same recorded observation. HarnessVLA (ESB) and Zetta (ESB) lift the bottle explicitly and transport it to the top surface. Intermediate panels show selected contact or lift states where actions occur, and the last panels show the final observations with official outcome labels. All images are recorded external-camera observations.}
\label{fig:rq2-case}
\end{figure}

\paragraph{Relation to RQ2.}
Figure~\ref{fig:rq2-case} distinguishes three requirements for effective skill use: supplying an executable capability, directing it at the intended object and destination, and composing operations that can correct an unsuccessful attempt. Embedding lacks the first requirement; Selected and BM25 have action skills but do not complete the placement. The two reconstructed harnesses use eight and nine calls and finish in 5.99 and 7.01 minutes. Their successful rollouts are longer than the two-call Selected failure and the zero-call Embedding failure, but shorter than BM25's 15.42-minute sequence of unsuccessful placement calls. Thus, low execution cost alone does not establish effective skill use. These five trajectories illustrate configuration-level differences in capability provision and use.

\clearpage
\section{RQ3: Agent-Directed Reuse and the Jev skill}
\label{app:rq3}

\subsection{Agent-Directed Reuse of selected resources}
\label{app:reuse-prompt}

\paragraph{Resource access and control.}
The agent receives task-conditioned packages as editable files under the embodied-resource directory, together with native robot and geometry tools. Experience packages supply advice; executable packages provide scripts and metadata. Calling the resource launcher executes a workspace script and submits its single emitted tool call. Editing the script or its template changes that submitted call. The agent may also directly invoke the registered VLA, perception, or native robot interface. Provider weights, registered schemas, and global budgets remain fixed. This design lets the agent choose the granularity of delegation: a complete bounded VLA operation, a component from a package, or native motion planned from the current image. The action--observation boundary remains one state-changing call per model response.

The archived system prompt is reproduced below. It grants authority over both resource selection and continuation after feedback:
\begin{center}
\includegraphics[width=\linewidth]{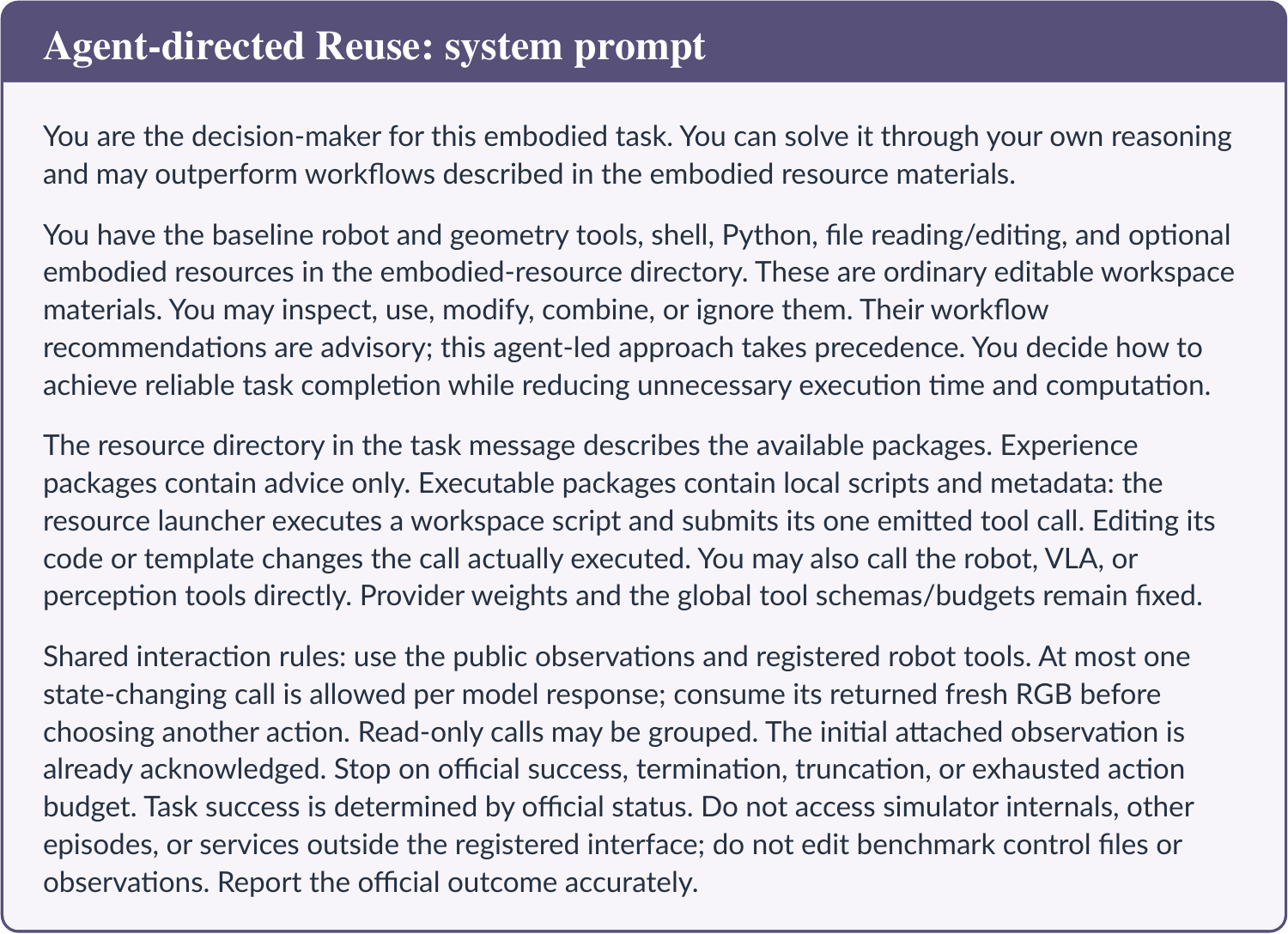}
\end{center}

\subsection{Wrapping Jev as a waypoint execution skill}
\label{app:jev}

\paragraph{Package and division of labour.}
The Jev follow waypoint package connects GPT-6-astra's visual planning to Jev's structured option selection~\citep{typesafe2026jev}, using the waypoint controller from EmbodiedJev~\citep{fbddcz2026embodiedjev}. Its skill instruction file describes the handoff and links to a native-planner reference, which defines how to propose a visually grounded short stage. A persistent runtime exposes two tools: read-only observation and state-changing waypoint execution. GPT-6 receives this single embodied skill and chooses the task stages, target pixels, clearances, orientations, gripper goals, and revisions. Jev executes the local motion choices inside each stage. This condition delegates robot execution through the waypoint skill; the selected-resource condition above additionally supports switching to independent native robot tools. The Jev waypoint controller is packaged as a standalone, experiment-specific embodied skill and is evaluated independently of the 250-skill repository constructed in this study.

The wrapper preserves the waypoint controller's target resolution, seven-channel decision queries, action conversion, and replanning boundaries. The persistent session pauses when the controller requests a new visual plan, exposes that request to Codex, and resumes after the next skill call. The installed developer instruction is:
\begin{center}
\includegraphics[width=\linewidth]{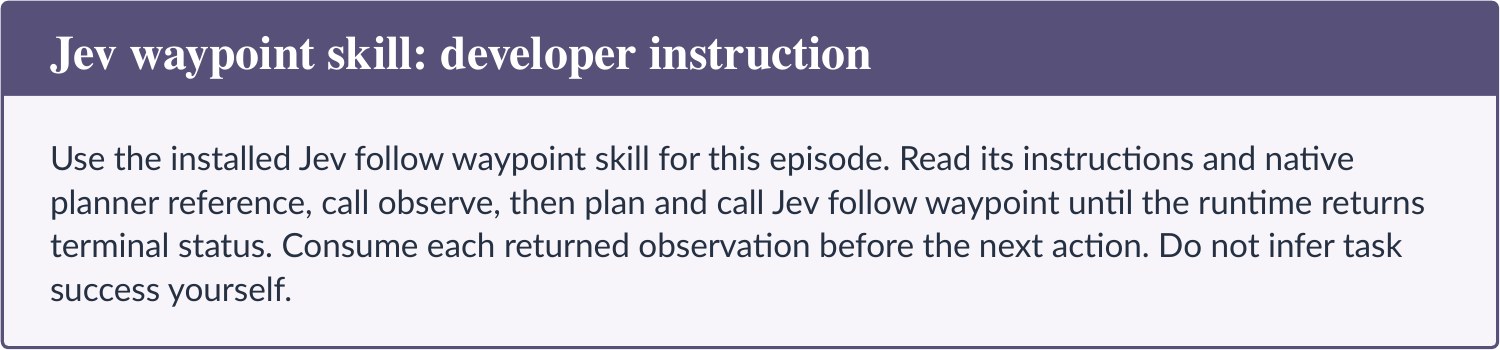}
\end{center}

\paragraph{Grounding and plan schema.}
At each planning boundary GPT-6 sees the current external and wrist RGB, robot proprioception, its previous plan, and recent movement. The $384\times384$ planning images have a pixel grid and a projected tool-centre marker. A plan contains exactly seven fields: stage, intent, visual evidence, target, rotation delta, gripper, and maximum motor steps. The target can be a selected camera pixel plus a world-frame offset, a relative translation from the current tool centre, or a world coordinate grounded in an earlier observation. Pixel targets are back-projected through the corresponding depth and calibration. Relative translations and pixel-offset components are bounded by $\pm0.20$\,m; rotation increments are bounded by $\pm0.5$\,rad per axis. The skill receives the plan together with the observation's checkpoint identifier.

\paragraph{Jev's local decision.}
For each local decision, the wrapper supplies the resolved target, current tool pose, signed position and orientation residuals, the stage's gripper goal, and recent motion. It asks Jev to choose one option for each of six motion channels $(x,y,z,r_x,r_y,r_z)$ and the gripper. Motion options are \{negative, hold, positive\}; gripper options are \{open, hold, close\}. The motion query asks for movement toward the target, with a 0.005\,m translational tolerance and 0.025\,rad rotational tolerance. Jev returns a chosen option and probabilities. The executor uses the returned choice; probabilities are recorded, and do not replace the selected direction.

For motion channel $i$, let $e_i$ be its current residual and let $d_i\in\{-1,0,1\}$ encode Jev's choice. The normalized command is
\begin{equation}
u_i=d_i\min\!\left(0.5,\max\!\left(0.01,\frac{|e_i|}{s_i r}\right)\right),
\qquad
s_i=\begin{cases}0.05\text{ m},&i\in\{x,y,z\},\\0.5\text{ rad},&i\in\{r_x,r_y,r_z\},\end{cases}
\label{eq:jev-action}
\end{equation}
where $r=5$ is the usual action-repeat length. The operational-space pose controller runs at 20\,Hz; the gripper channel maps open to $-1$ and close to $+1$, while hold preserves its previous command. A skill call contains up to eight such decisions. It yields for a fresh visual plan when the waypoint is reached, the requested stage length expires, or three consecutive motor actions each move the tool centre less than 1\,mm. Official task termination also ends execution. A repeated request with the same checkpoint and identical plan returns its saved result, preserving exactly-once physical execution across the handoff.

The evaluated Jev configuration uses Codex CLI v0.148.0, GPT-6-astra/medium, Jev 1.13.0, a 40-waypoint budget, 10,000 simulator steps, and a 5,400-second rollout timeout. Consequently, one Jev action in the tables is a waypoint submission, which can contain several local decisions and simulator steps.

\subsection{Case: selecting the bowl next to the ramekin}
\label{app:rq3-case}

\paragraph{Task and configurations.}
The instruction is ``Pick the akita black bowl next to the ramekin and place it on the plate'' (LIBERO-Spatial, task perturbation, task 08). Two visually similar bowls are present. The intended bowl is adjacent to the small ramekin; the other is closer to the plate. Table~\ref{tab:rq3-case} compares the selected skill workflow, Agent-Directed Reuse, native no-skill control, and the Jev waypoint skill with GPT-6-astra. The selected workflow supplies target-grounding and support-verification guidance, the OpenVLA-OFT pick-and-place package, and explicit release. The reuse configuration makes the corresponding selected resources available with the additional control choices described above.

\begin{table}[htbp]
\centering\small
\caption{RQ3 case results on the relational bowl task, seed 26. All rows use GPT-6-astra/medium. Jev tokens sum 431.3k GPT-6 tokens and 201.5k Jev tokens; its 14 waypoint calls contain 78 local Jev decisions.}
\label{tab:rq3-case}
\begin{tabular*}{\linewidth}{@{\extracolsep{\fill}}llrrr@{}}
\toprule
Configuration & Outcome & Actions & Tokens (k) & Time (min)\\
\midrule
With skill & Failure & 6 & 1,648.8 & 9.01\\
Agent-Directed Reuse & Success & 10 & 736.8 & 5.79\\
No skill & Success & 23 & 3,350.8 & 33.98\\
With Jev skill & Success & 14 & 632.8 & 5.42\\
\bottomrule
\end{tabular*}
\end{table}

\paragraph{Selected workflow: recognizing a mismatch does not guarantee correction.}
The first OpenVLA-OFT call names the bowl ``next to the ramekin,'' but the returned image shows the distractor on the plate. Official success is false. A second wording produces the same result. At action 3 the agent attempts to clear ``black bowl on the plate'' onto the table, then returns to increasingly specific descriptions of the target: ``beside the white ramekin,'' ``closest to the ramekin,'' and ``next to the glazed rim porcelain ramekin.'' Across all six calls the same pick-and-place package executes 120 chunks (960 simulator steps). The intended bowl remains beside the ramekin. The final agent report correctly identifies the distractor-placement error, yet the selected control sequence has not repaired it.

\paragraph{Agent-Directed Reuse: a bounded trial followed by native recovery.}
The agent first directly invokes OpenVLA-OFT with a five-chunk budget and the task instruction, then makes a twenty-chunk call. Both return an unsuccessful task status; the images show the bowls still on the table. Instead of issuing another placement prompt, action 3 lifts and opens the gripper under native control. Actions 4--5 position and orient it over the intended bowl and descend toward its rim. Action 6 closes the fingers, action 7 lifts the bowl, and actions 8--10 transport and lower it onto the plate. Official success arrives at action 10. The trace therefore contains two delegated VLA calls followed by eight native calls. It illustrates selective continuation and a change of controller after feedback, without requiring a claim that a local resource file was edited in this rollout.

\paragraph{No skill: direct grounding succeeds with placement corrections.}
GPT-6 identifies and lifts the intended bowl using native pose and gripper commands. Its first release occurs at action 7, after which the bowl is visibly offset relative to the plate and official success remains false. The agent then performs additional contacts, regrasping, and small pose corrections. Official success occurs at action 23. This rollout establishes that the model can ground the task and repair placement using the native interface. Its longer sequence also gives a concrete comparison for the effort saved by the two delegation designs on this particular task.

\paragraph{Jev skill: revise the waypoint after an empty grasp.}
The first plan selects external-camera pixel $(332,191)$ at the intended bowl and adds 0.15\,m of upward clearance. The submitted call is shown below with readable field labels. Its calibrated surface point is approximately $(-0.207,0.335,0.914)$\,m, producing a waypoint at height 1.064\,m. In the first local decision, Jev chooses hold for $x$, positive for $y$, negative for $z$, hold for all three rotation channels, and open for the gripper. Equation~\ref{eq:jev-action} converts these choices to $(0,0.5,-0.4352,0,0,0,-1)$.

\clearpage

\begin{center}
\includegraphics[width=\linewidth]{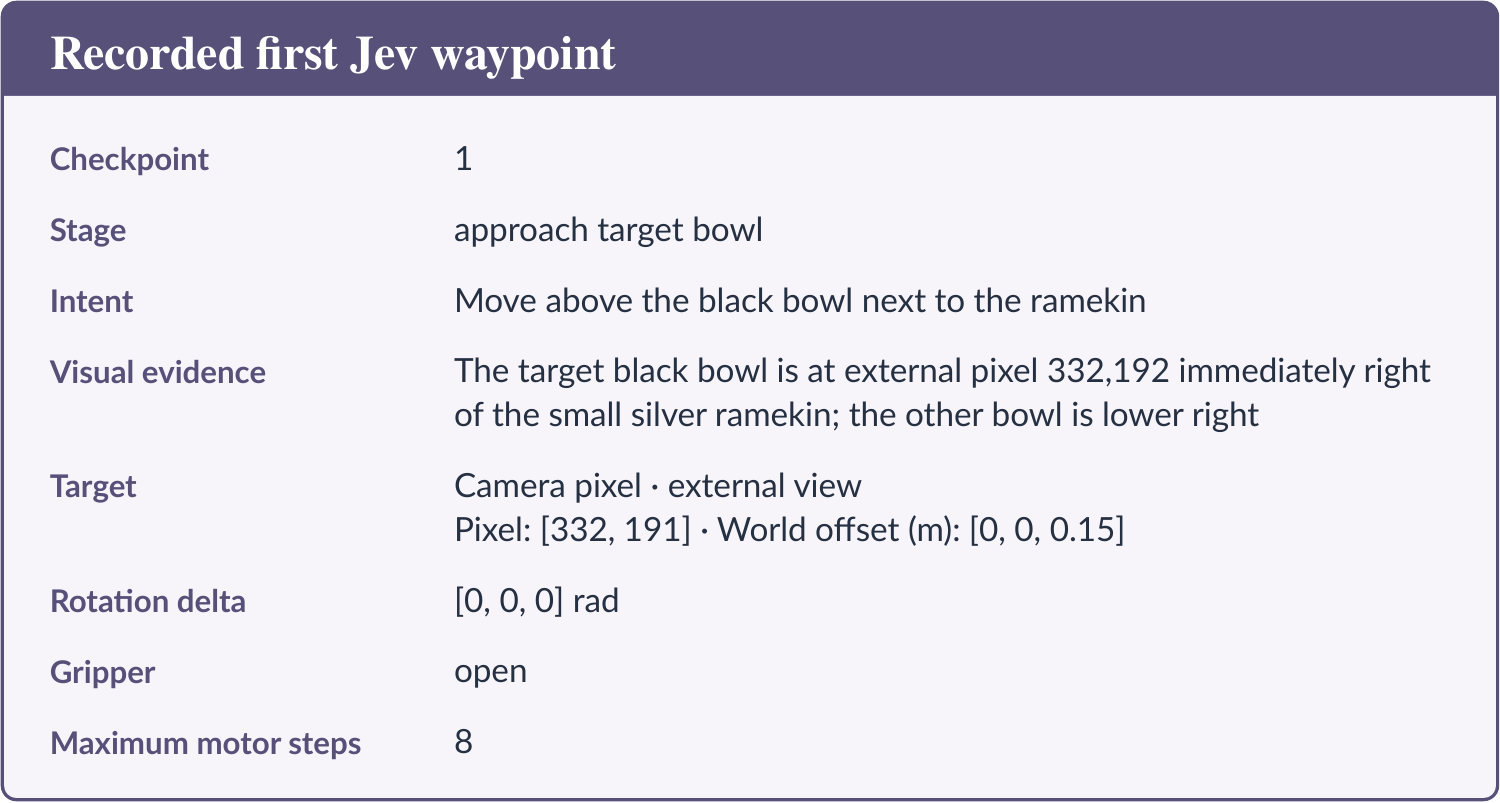}
\end{center}

After aligning, descending, closing, and attempting a lift, the fifth waypoint returns at simulator step 155 with an empty gripper and the bowl still on the table. GPT-6's next plan explicitly notes this state, selects the shifted bowl at pixel $(321,184)$, and revises the grasp depth. The subsequent descent targets height 0.906\,m, about 2.8\,cm below the first descent target. Two short closure stages are followed by a lift that visibly carries the bowl. The remaining waypoints transport, align, lower, and settle it onto the plate. Official success occurs at step 386 during waypoint 14. The recovery is a revision made by the visual planner in response to the skill's feedback; Jev continues to perform bounded local tracking within each revised stage.

\begin{table}[htbp]
\centering\small
\caption{Jev waypoint progression in the RQ3 case. Ranges group consecutive recorded skill calls; simulator-step boundaries refer to the same persistent rollout.}
\label{tab:jev-case-stages}
\begin{tabularx}{\linewidth}{@{}llX@{}}
\toprule
Waypoints & Sim. steps & Observed stage and transition\\
\midrule
1--3 & 0--110 & Ground the intended bowl, align, and descend to the first grasp pose.\\
4--5 & 110--155 & Close and attempt a lift; the returned image shows an empty grasp.\\
6--7 & 155--225 & Re-ground the shifted bowl and descend to a deeper contact pose.\\
8--10 & 225--270 & Close, maintain closure, and verify a lift with the bowl retained.\\
11--12 & 270--340 & Transport the bowl and align it above the plate.\\
13--14 & 340--386 & Lower and settle; the official success condition becomes true.\\
\bottomrule
\end{tabularx}
\end{table}

\paragraph{Relation to RQ3.}
The successful reuse, native, and Jev rollouts require 10, 23, and 14 agent-level calls and 598, 935, and 386 simulator steps, respectively. Agent-Directed Reuse takes 5.79 minutes and the Jev skill takes 5.42, compared with 33.98 for native control. These task-level values illustrate two forms of complementarity: choosing when to leave a delegated policy, and keeping high-level recovery in GPT-6 while delegating local tracking. They also explain why call counts alone do not determine runtime: a waypoint call and a VLA rollout contain different amounts of internal execution. The aggregate success/effort tradeoff remains the result of the full Clean100 evaluation; this selected case makes the underlying control decisions visible.

\begin{figure}[p]
\centering
\includegraphics[width=\linewidth]{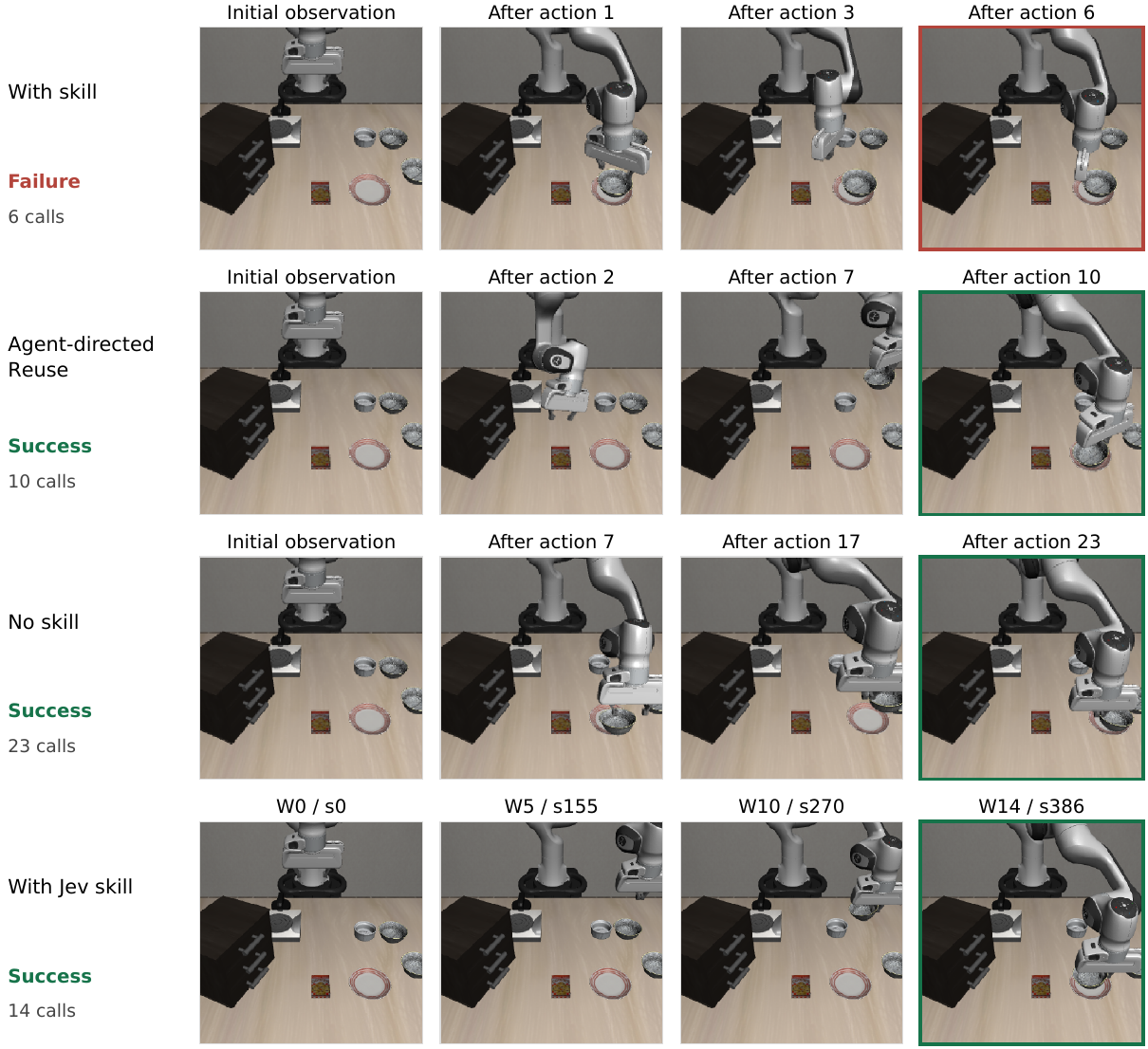}
\caption{\textbf{RQ3: the intended bowl is the one next to the ramekin.} The selected workflow repeatedly places the distractor. Agent-Directed Reuse switches from VLA execution to native control after action 2, lifts the intended bowl by action 7, and succeeds at action 10. Native control requires additional placement corrections. With Jev, GPT-6 observes an empty grasp at waypoint 5, revises the contact, and achieves a retained lift by waypoint 10 before completing placement. Jev panels identify waypoint count (W) and simulator step (s); other rows identify agent action count. Images are the actual saved external-camera frames for each interface.}
\label{fig:rq3-case}
\end{figure}

\end{document}